\documentclass[12pt]{iopart}

\usepackage{iopams}  

\usepackage{graphicx}
\usepackage{color}
\usepackage{hyperref}
\hypersetup{colorlinks=true,citecolor=magenta,linkcolor=blue}
\usepackage{tikz}
\usetikzlibrary{arrows,shapes}
\usepackage{adjustbox}
\usepackage{tikz}
\usetikzlibrary{arrows,shapes}
\usetikzlibrary{shapes.multipart}
\usetikzlibrary{arrows,positioning}
\tikzset{
    >=stealth',
    punkt8/.style={
           rectangle,
           rounded corners,
           draw=black, very thick,
           text width=8em,
           minimum height=2em,
           text centered},
    pil/.style={
           ->,
           thick,
           shorten <=2pt,
           shorten >=2pt,}
}

\tikzset{
    >=stealth',
    punkt6/.style={
           rectangle,
           rounded corners,
           draw=black, very thick,
           text width=6em,
           minimum height=2em,
           text centered},
    pil/.style={
           ->,
           thick,
           shorten <=2pt,
           shorten >=2pt,}
}

\tikzset{
    >=stealth',
    punkt4/.style={
           rectangle,
           rounded corners,
           draw=black, very thick,
           text width=4em,
           minimum height=2em,
           text centered},
    pil/.style={
           ->,
           thick,
           shorten <=2pt,
           shorten >=2pt,}
}

\tikzset{
    >=stealth',
    punkt12/.style={
           rectangle,
           rounded corners,
           draw=black, very thick,
           text width=12em,
           minimum height=2em,
           text centered},
    pil/.style={
           ->,
           thick,
           shorten <=2pt,
           shorten >=2pt,}
}
\begin{document}

\title[A multiscale and multicriteria GAN to synthesize 1-dimensional turbulent fields.]{A multiscale and multicriteria Generative Adversarial Network to synthesize 1-dimensional turbulent fields}

\author{Carlos Granero Belinchon$^{1,2}$ \& Manuel Cabeza Gallucci$^{1, 3}$}
\address{$^{1}$ Department of Mathematical and Electrical Engineering, IMT Atlantique, Lab-STICC, UMR CNRS 6285, 655 Av. du Technop\^ole, Plouzan\'e, 29280, Bretagne, France. \\
$^{2}$ Odyssey, Inria/IMT Atlantique, 263 Av. G\'en\'eral Leclerc, Rennes, 35042, Bretagne, France. \\
$^{3}$ Electronic Engineering Department, Facultad de Ingenieria, Universidad de Buenos Aires, Paseo Colon 850, C1063ACV, Buenos Aires, Argentina}
\ead{carlos.granero-belinchon@imt-atlantique.fr ; mcabezag@fi.uba.ar}

\vspace{10pt}
\begin{indented}
\item[]August 2023
\end{indented}

\begin{abstract}
This article introduces a new Neural Network stochastic model to generate a 1-dimensional stochastic field with turbulent velocity statistics. Both the model architecture and training procedure ground on the Kolmogorov and Obukhov statistical theories of fully developed turbulence, so guaranteeing descriptions of 1) energy distribution, 2) energy cascade and 3) intermittency across scales in agreement with experimental observations. The model is a Generative Adversarial Network with multiple multiscale optimization criteria. First, we use three physics-based criteria: the variance, skewness and flatness of the increments of the generated field, that retrieve respectively the turbulent energy distribution, energy cascade and intermittency across scales. Second, the Generative Adversarial Network criterion, based on reproducing statistical distributions, is used on segments of different length of the generated field. Furthermore, to mimic multiscale decompositions frequently used in turbulence's studies, the model architecture is fully convolutional with kernel sizes varying along the multiple layers of the model. To train our model, we use turbulent velocity signals from grid turbulence at Modane wind tunnel.
\end{abstract}
\vspace{2pc}
\noindent{\it Keywords}: TURBULENCE, STOCHASTIC FIELDS, NEURAL NETWORK , GAN
%\submitto{\jpa}
%\maketitle

\section{Introduction}

Turbulent fluids exhibit complex non-linear and multiscale dynamics which can not be described from a deterministic point of view and which lead to a complex statistical behavior of the velocity field of the flow~\cite{Kolmogorov1991,Kolmogorov1962,Obukhov1962,Frisch1995}. Therefore, not only second-order but also higher-order statistics are needed to describe turbulent velocity fields~\cite{Atta1980, Anselmet1984, Gagne1990, Meneveau1991, Lohse1993, Frisch1995}. The generation of stochastic fields with turbulent statistics have been broadly studied~\cite{Flandrin1989,Flandrin1992,Benzi1993,Arneodo1998,Bacry2001,Robert2008,Chevillard2012} and still remain a subject of study~\cite{Du2018,Chevillard2019, Peinke2019, Alexandrov2020}. Several stochastic models based on the multifractal description of turbulence have been developed in the last decades. Among them, the fractional Brownian motion only recovering second-order statistics~\cite{Mandelbrot1968,Flandrin1989,Flandrin1992}, and more complex models recovering second and higher-order ones~\cite{Benzi1993,Arneodo1998,Bacry2001,Chevillard2012}. Some of these models are discrete in the scales~\cite{Benzi1993,Arneodo1998} while others are continuous~\cite{Bacry2001,Chevillard2012}. Stochastic models of turbulence based on the Focker-Planck equation were also developed. Some of them only recovering second-order statistics~\cite{Peinke1993} and more recent ones reproducing also higher-order statistics~\cite{Nawroth2006}. Other approaches are based on linear filtering~\cite{Klein2003, Hoepffner2011} and spectral tensor models~\cite{Mann1998}, but they only recover second-order statistics.

In the last decade, Neural Network (NN) stochastic generative models appeared and gained popularity, from Generative Adversarial Networks (GANs)~\cite{Goodfellow2014,Beroud2023}, to generative diffusion models~\cite{Song2019,Yan2021,Dhariwal2021} or autoencoders~\cite{Zhang2020,Ye2022}. These models are being currently developed to generate stochastic fields with complex statistical structures~\cite{Beroud2023} and more particularly with turbulent statistical behavior~\cite{Brunton2020,Beck2021,Li2023a,Li2023b}. Thus, NN stochastic generative models have been used to reproduce turbulent fields for different applications~\cite{Deng2019,Liu2020,Kim2020,Kim2021,Geneva2020,Drygala2022,Wang2020,Buzzicotti2021}. In order to generate satisfactory stochastic fields, these NN approaches need the most of the time to incorporate some physics of turbulence~\cite{Kim2019,Wu2020,Yousif2021,Yousif2022,Yousif2022a}. However, very few of these works incorporate higher-order statistics in the training procedure~\cite{Wu2020}, and the most only evaluate the performance of the NN model to reproduce second-order statistics of the field, mainly the power spectrum or the correlation function~\cite{Liu2020,Yousif2021,Yousif2022,Kim2020}. Then, by not evaluating higher-order statistics of the generated field, these works do not study the capacity of their models to reproduce the energy cascade of turbulence~\cite{Kolmogorov1991} or the intermittency phenomenon~\cite{Kolmogorov1962,Obukhov1962}.

In this article, we propose a multiscale and multicriteria GAN to generate a 1-dimensional stochastic field with turbulent statistics. This model grounds on the Kolmogorov and Obukhov theories of turbulence~\cite{Kolmogorov1991,Kolmogorov1962,Obukhov1962}, and so, contrary to most of the state of the art, it includes second-order and higher-order statistics in both training and evaluation. Consequently, the proposed model is based on the physics of turbulence, and the generated stochastic field correctly reproduces the energy distribution, energy cascade and intermittency of turbulent velocity flows. From the best of our knowledge, this work, together with~\cite{GraneroBelinchon2024}, are the first Neural Network generative models directly grounding on the Kolmogorov-Obukhov theories. Moreover, the GAN approach used in this work allows to include generative criteria based on the full probability density function (PDF) of turbulent velocity and so to overcome some limitations of~\cite{GraneroBelinchon2024}.

Section~\ref{sec:turb} presents the Kolmogorov-Obukhov theories of fully developed turbulence and the experimental velocity dataset used for training. Section~\ref{sec:model} introduces the multiscale and multicriteria physics-based GAN approach used to generate a 1d stochastic field with turbulent statistics. Section~\ref{sec:results} shows a second and higher-order statistical description of the generated field and the experimental velocity dataset and compares both statistical behaviors. Finally, section~\ref{sec:conclusions} presents the main conclusions and perspectives.

\section{Fully developed turbulence} \label{sec:turb}

\subsection{The Kolmogorov-Obukhov theories: energy cascade and intermittency}

Fully developed turbulence corresponds to flows at very high Reynolds number, when the nonlinear advection term of the Navier-Stokes equations dominates the dynamics of the flow~\cite{Frisch1995}. In this limit and away from boundaries, the flow is considered statistically isotropic and homogeneous and the Kolmogorov-Obukhov theories provide a statistical description of fully developed turbulence~\cite{Kolmogorov1991,Kolmogorov1962,Obukhov1962}.

The Kolmogorov 1941 theory prescribes the existence of three ranges of scales with different statistical behaviors: the integral domain contains the large scales of the flow, where energy is injected, the dissipative domain contains the small scales, where energy is dissipated, and the inertial domain contains the scales in between, where the energy cascades from large scales down to smaller ones~\cite{Frisch1995,Kolmogorov1991, Richardson1921}. Thus, two scales appear naturally: the integral scale $L$ which divides the integral and inertial domains, and the Kolmogorov scale $\eta$ which separates the inertial and dissipative ones.

Following Kolmogorov 1941, we focus on the longitudinal velocity $v(x)$~\cite{Kolmogorov1991,Frisch1995}, and we use the increment to define the scale $l$ of turbulent velocity:

\begin{equation}
\delta_l v(x)=v(x+l)-v(x)
\end{equation}

In the inertial domain of scales $L>l>\eta$, the Kolmogorov 1941 theory states that the structure function of order $p$, defined as the statistical moment of order $p$ of the velocity increments, behaves as a power law of the scale with exponent $p/3$:

\begin{equation} \label{eq:K41}
 S_p(l)=\left\langle \left( \delta_l v(x) \right) ^p \right\rangle \propto l^{p/3}
\end{equation}

\noindent where $S_p(l)$ is the structure function of order $p$ and $\left\langle \right\rangle$ is the spatial average. 

The variance of the velocity increments, $S_2(l)$, describes the energy distribution of the flow across scales, and from (\ref{eq:K41}), it behaves as $l^{2/3}$ in the inertial domain. Moreover, the $4/5$ law of Kolmogorov, directly obtained from the Navier-Stokes equation through the Karman-Howarth development~\cite{Karman1938}, shows the existence of an energy flux in the inertial domain from large to small scales. This energy flux is related to the third-order statistical moment of the velocity increments $S_3(l) \propto -l$, and so, the $4/5$ law of Kolmogorov directly relates the energy cascade across scales to the non-Gaussianity of the velocity field.

Finally, Kolmogorov and Obukhov provided a correction to the above theory in 1962~\cite{Kolmogorov1962,Obukhov1962}: the energy dissipation in a turbulent flow is non homogeneous and should be considered locally. This leads to a deformation of the PDF of the velocity increments across the scales, from almost Gaussian at large scale to non-Gaussian at small scales~\cite{Anselmet1984,Meneveau1991,Lohse1993,Chevillard2005,Chevillard2012}. More precisely, extreme events of the velocity increments are the more and more intense and the more and more recurrent at small scales. This is known as intermittency, and leads to the following correction to the Kolmogorov 1941 theory:

\begin{equation} \label{eq:Ko62}
  S_p(l)=\left\langle \left( \delta_l v(x) \right) ^p \right\rangle \propto l^{\zeta_{p}}
\end{equation}

\noindent where $\zeta_{p}$ is the scaling exponent which is a non-linear function of $p$. From the $4/5$ law of Kolmogorov $\zeta_{3}=1$. The non-linear scaling exponent can be measured by fitting the slope of $\log(S_p(l))$ in function of $\log(l)$ in the inertial domain of scales for different orders $p$.

In order to highlight the non-Gaussian and intermittent nature of turbulence, we focus on the skewness and flatness of the velocity increments~\cite{Atta1980, Gagne1990, Frisch1995, Tabeling1996, Chevillard2010} defined as:

\begin{eqnarray}
    \mathcal{S}(l)= \frac{S_3(l)}{S_2(l)^{3/2}} =\frac{\left\langle \left( \delta_l v(x) \right) ^3 \right\rangle}{\left( \left\langle \delta_l v(x) \right) ^2 \right\rangle ^{3/2}} \label{eq:skewness}\\
    \mathcal{F}(l)= \frac{S_4(l)}{S_2(l)^{2}} = \frac{\left\langle \left( \delta_l v(x) \right) ^4 \right\rangle}{\left( \left\langle \delta_l v(x) \right) ^2 \right\rangle ^{2}} \label{eq:flatness}
\end{eqnarray}

On the one hand, the skewness characterizes the assymmetry of the PDF of the velocity increments across scales. Symmetric PDFs exhibit zero skewness ($\mathcal{S}=0$), and from the Kolmogorov 4/5 law, negative values of the skewness in the inertial and dissipative domains reflect the existence of an energy cascade. On the other hand, the flatness characterizes the importance of the tails of the PDF of the velocity increments across scales. A Gaussian PDF is characterized by $\mathcal{F}=3$, and so, the flatness of the velocity increments is $3$ in the integral domain of scales and increases to non-Gaussian values when the scale decreases. This illustrates the higher importance of extreme events in the PDF of the velocity increments at small scales~\cite{Moffatt2021}, \textit{i.e.} the intermittent nature of turbulence.  

\subsection{Modane turbulent velocity dataset}

The Modane turbulent velocity dataset consists on Eulerian longitudinal velocity measurements from a grid turbulence setup in the wind tunnel of ONERA at Modane~\cite{Kahalerras1998}. The mean velocity of the flow is $\left\langle v \right\rangle = 20.5 \,\, m/s$. The measurements are obtained at a sampling frequency of $f_s=25 \,\, kHz$ at a fixed point far away from boundaries. The Taylor-scale Reynolds number of the flow is $R_{\lambda}=2500$. Then, we consider the flow in fully developed turbulent regime as well as homogeneous and isotropic. We assume Taylor frozen turbulence hypothesis and consider temporal variations as spatial ones~\cite{Frisch1995}. Then, the sampling distance can be expressed as $l_s=\left\langle v \right\rangle f_s$. From previous studies, the integral and Kolmogorov scales of the flow are $L=2350 \, l_s$ and $\eta=5 \, l_s$~\cite{GBelinchon2016}. A detailed multiscale statistical characterization of the Modane turbulent velocity dataset is provided in section~\ref{sec:results} and in the literature~\cite{Gagne1990,Kahalerras1998,Chevillard2012,Arneodo1999}. 

\section{Multiscale and Multicriteria Physics Based GAN} \label{sec:model}

Our Neural Network model is conceived to synthesize a stochastic field $u(x)$ reproducing four main results of the Kolmogorov-Obukhov theory of fully developed turbulence:

\begin{itemize}
    \item In a turbulent flow, we can discriminate \textbf{three domains of scales}: integral, inertial and dissipative, each one with a different statistical behavior.
    \item The \textbf{distribution of energy} across scales is described by $S_2(l)$, and in the inertial domain it behaves as $\sim l^{2/3}$ up to corrections due to intermittency.
    \item The \textbf{cascade of energy} is related to the non-Gaussianity of the velocity field through the $4/5$ law of Kolmogorov. This implies negative values of the skewness in both the inertial and dissipative domains, as well as $\zeta_3=1$.
    \item \textbf{Intermittency} introduces extreme events in the velocity increments at scales in the inertial and dissipative domains. The importance of these extreme events is characterized by the flatness, which tends to Gaussian values at large scales. Intermittency also leads to a non-linear scaling exponent function $\zeta_p$.
\end{itemize}

\begin{figure}[!ht]
\begin{center}
\begin{adjustbox}{max totalsize={\textwidth}{.7\textheight},center}
\begin{tikzpicture}[node distance=1cm, auto, every text node part/.style={align=center}]
 \node[punkt6, scale=1] (1) at (-4.8,1) {Gaussian noise $w(x)$};
 \node[punkt8, scale=1, color=red] (2) at (-4.8,-1) {$u(x)=\mathcal{G}(w(x))$};
 \node[punkt6, scale=1] (3) at (-1,1) {Modane samples $v(x)$};
 \node[punkt6, scale=1] (4) at (-1,-1) {Generated samples $u(x)$};
 \node[punkt8, scale=1, color=blue] (5) at (3,2) {$\mathcal{D}_{\textrm{scale-invariance}}$};
 \node[punkt4, scale=1, color=blue] (6) at (3,0.66) {$\mathcal{D}_{S_2}$};
 \node[punkt4, scale=1, color=blue] (7) at (3,-0.66) {$\mathcal{D}_{\mathcal{S}}$};
 \node[punkt4, scale=1, color=blue] (8) at (3,-2) {$\mathcal{D}_{\mathcal{F}}$};
 \node[punkt8, scale=1, color=green!40!black] (9) at (7.5,2) {Scale-invariance loss $l_{SI}$};
 \node[punkt6, scale=1, color=green!40!black] (10) at (7.5,0.66) {$l_{S_2}$ loss};
 \node[punkt6, scale=1, color=green!40!black] (11) at (7.5,-0.66) {$l_{\mathcal{S}}$ loss};
 \node[punkt6, scale=1, color=green!40!black] (12) at (7.5,-2) {$l_{\mathcal{F}}$ loss};
 \node[punkt12, scale=1, color=green!40!black] (13) at (7.5,-3.9) {$\mathcal{L} = \alpha l_{SI} + \beta l_{S_2} + \gamma l_{\mathcal{S}} + \lambda l_{\mathcal{F}}$ };

 \draw[color=blue!70!black,ultra thick] (1,3) rectangle (5,-3);
 \draw[color=green!50!black,ultra thick] (5.4,3) rectangle (9.6,-3);
 %\draw[color=black,ultra thick] (10.8,0) circle (12pt); 
 %\draw (10.8,0) node {\Large{+}};

 \path[->]
%        FROM                        BEND/LOOP                                POSITION OF LABEL                        LABEL        TO
         (1)                edge[color=black!70!black,pil,bend left=0]        node[swap, very near start]        {}                (2)
         (2)                edge[color=black!70!black,pil,bend left=0]        node[swap, very near start]        {}                (4)
         (3)                edge[color=black!70!black,pil,bend left=0]        node[swap, very near start]        {}                (1,1)
         (4)                edge[color=black!70!black,pil,bend left=0]        node[swap, very near start]        {}                (1,-1)
         (5)                edge[color=black!70!black,pil,bend left=0]        node[swap, very near start]        {}                (9)
         (6)                edge[color=black!70!black,pil,bend left=0]        node[swap, very near start]        {}                (10)
         (7)                edge[color=black!70!black,pil,bend left=0]        node[swap, very near start]        {}                (11)
         (8)                edge[color=black!70!black,pil,bend left=0]        node[swap, very near start]        {}                (12)
         (7.5,-3)                edge[color=black!70!black,pil,bend left=0]        node[swap, very near start]        {}                (7.5,-3.5);
        % (9.3,2)                edge[color=black!70!black,pil,bend left=0]        node[swap, very near start]        {}                (10.4,0)
        % (8.9,0.66)                edge[color=black!70!black,pil,bend left=0]        node[swap, very near start]        {}                (10.4,0)
        % (8.9,-0.66)                edge[color=black!70!black,pil,bend left=0]        node[swap, very near start]        {}                (10.4,0)
        % (8.9,-2)                edge[color=black!70!black,pil,bend left=0]        node[swap, very near start]        {}                (10.4,0);
\end{tikzpicture}
\end{adjustbox}
\caption{Multiscale and multicriteria physics based GAN. In red the fully convolutional generator model $\mathcal{G}$, which produces realizations of a 1d stochastic field, $u(x)$, from realizations of a Gaussian noise, $w(x)$. In blue the physics-based discriminators used to train the model. The $\mathcal{D}_{\textrm{scale-invariance}}$ discriminator is fed with Modane realizations $v(x)$ and generated ones $u(x)$, while the other three discriminators are directly fed with the corresponding statistics across scales of $v(x)$ and $u(x)$. Each discriminator has its own loss function in green. The total loss function of the GAN, also in green, is a linear combination of the four loss functions of the discriminators.}\label{fig:model}
\end{center}
\end{figure}
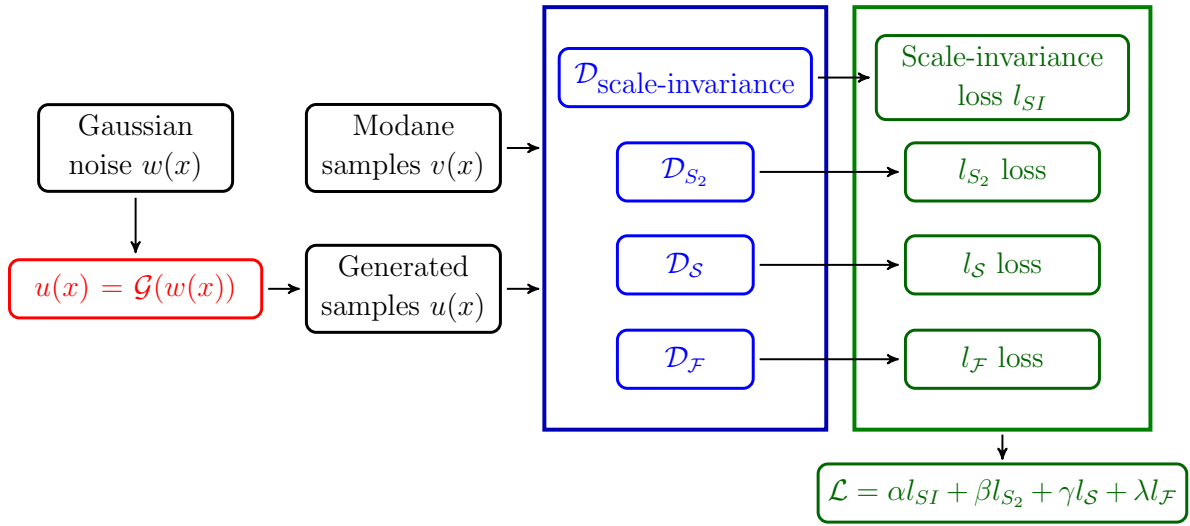

Our model is based on the GAN approach first proposed by Goodfellow et al.~\cite{Goodfellow2014}. However, several modifications have been done to introduce the physics of turbulence into the model. First, the generator model $\mathcal{G}$ follows a fully-convolutional U-Net structure which, similarly to the Kolmogorov-Obukhov theory of turbulence, is based on multiresolution analysis. Second, several multiscale criteria are used during training, each one with an assigned discriminator network. Thus, in order to recover respectively the energy distribution, energy cascade and intermittency of turbulence, three discriminators, $\mathcal{D}_{S_2}$, $\mathcal{D}_{\mathcal{S}}$ and $\mathcal{D}_{\mathcal{F}}$, compare the $\log(S_2(l))$, $\mathcal{S}(l)$ and $\log(\mathcal{F}/3)$ of the generated field and Modane. These three criteria are also helpful for recovering the three domains of scales of turbulence. Finally, a discriminator, $\mathcal{D}_{\textrm{\small{scale-invariance}}}$, compares segments of different sizes of the generated field $u(x)$ and Modane $v(x)$. This criterion exploits the potentialities of GAN to reproduce the PDF of Modane data on segments of different sizes. It is devoted to impose turbulent statistical dynamics on $u(x)$ at different scales, that is, turbulent dynamics on the full process containing several integral scales, but also on small segments up to sizes of the order of $L$.
Figure~\ref{fig:model} illustrates this multiscale and multicriteria physics based Generative Adversarial Network approach.

\subsection{The generator model}\label{sec:generator}

The generator model $\mathcal{G}$ is fully-convolutional and stochastic:

\begin{equation}
u(x) = \mathcal{G}(w(x))
\end{equation}

\noindent This model takes as input a 1-dimensional Gaussian white noise, $w(x)$, of size $N$ and produces a 1-dimensional field of the same size, $u(x)$, with turbulent velocity statistics. Consequently, the model $\mathcal{G}$ operates doubly on the input Gaussian white noise. On the one hand, it deformates the Gaussian PDF of the noise to a PDF in agreement with turbulent velocity statistics, \textit{i.e.} slightly skewed. On the other hand, it introduces a structure of dependencies in the generated field leading to the desired statistical moments of the increments of the field.

\begin{figure}[!htb]
\centering
\includegraphics[width=\textwidth]{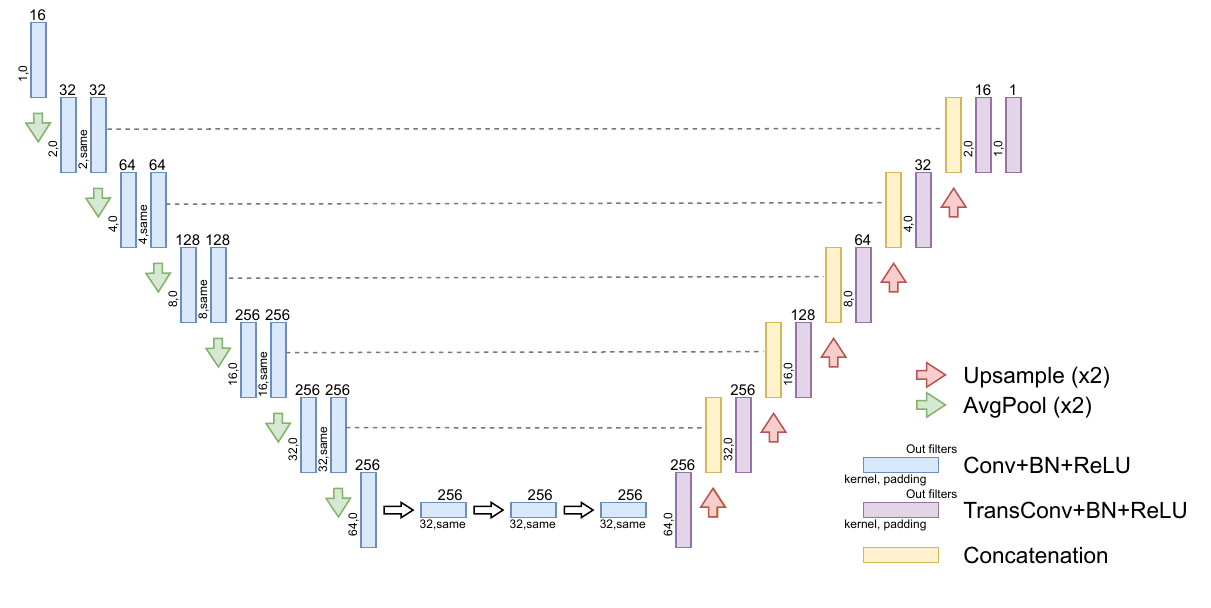}
\caption{U-Net architecture of $\mathcal{G}$ taking as input a Gaussian white noise, $w(x)$, of size $N$ and providing as output a stochastic field with turbulent statistics, $u(x)$, of the same size. Convolutional and transpose convolutional blocks, made up of a convolutional layer (respectively transpose convolutional layer), batch normalization and ReLU activation function, are represented by the blue and purple rectangles respectively. The number at the top of each rectangle indicates the number of channels of the output of the block. The number at the left of each rectangle indicates the kernel size of the filter. The used padding is indicated with \textit{same} when the size of the input and output are equal and $0$ when there is no padding. The stride used on each convolution and transpose convolution is always $1$. Green and red arrows mean respectively average pooling and upsampling. The concatenated long-skip connections are represented by the yellow rectangles.}
\label{fig:generator_model}
\end{figure}

The model $\mathcal{G}$ follows a U-Net architecture~\cite{Ronneberger2015}, which consists of an encoder followed by its symmetrical decoder both connected by a bridge at the deepest level, see figure~\ref{fig:generator_model}. On the one hand, our encoder has six levels, each one with two convolutional blocks and an average pooling layer. Each convolutional block is defined by a convolutional layer, batch normalization and ReLU activation function. On the other hand, the decoder has six analogous levels, each one defined by a transpose convolutional block (transpose convolution, batch normalization and ReLU activation) and an upsampling layer. The kernel sizes of the convolutions and transpose convolutions increase with the depth of the level and vary from 2 to 64 samples. This, together with the pooling, allows the model to reproduce the long range dependencies of the turbulent velocity field. The deepest level of the encoder and the decoder are bridged by three convolutional blocks, all with a kernel size of 32 samples. Moreover, concatenated long-skip connections link each level of the encoder with the symmetrical level of the decoder. This connections help in keeping the stability of the network during training and minimizing the vanishing gradient effects~\cite{Goodfellow-et-al-2016}. Concatenated long-skip connections were used since they provided the best performances in our case study. 
This generator is based on~\cite{GraneroBelinchon2024}, where the encoder and decoder were built mimicking the structure of dyadic wavelets decomposition and random wavelet cascade models of turbulence~\cite{Arneodo1998}. The dyadic evolution of the size of the convolution kernels, together with the poolings and upscalings, allows the generator to operate at scales from the dissipative to the integral domain of the flow.
The generator model has 26 millions of parameters.

\subsection{The multiscale and multicriteria discriminators}\label{sec:discriminators}

\begin{figure}[!htb]
\centering
\includegraphics[scale=0.8]{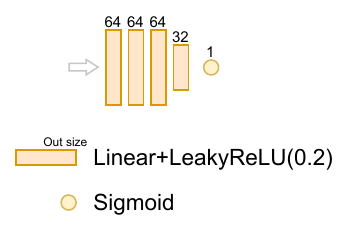}
\caption{Architecture of the discriminator models $\mathcal{D}_{S_2}$, $\mathcal{D}_{\mathcal{S}}$ and $\mathcal{D}_{\mathcal{F}}$, which take as input respectively the $\log(S_2(l))$, $\mathcal{S}(l)$ and $\log(\mathcal{F}/3)$ curves and provide a scalar value in $\left(0,1\right)$. Each yellow rectangle represents a dense layer followed by Leaky ReLU activation. The number at the top of each rectangle indicates the output size of the dense layer. All Leaky ReLu use a slope of $0.2$ for negative values. The yellow circle represents the sigmoid activation function.}
\label{fig:model_disc_structs}
\end{figure}

Three discriminators, $\mathcal{D}_{S_2}$, $\mathcal{D}_{\mathcal{S}}$ and $\mathcal{D}_{\mathcal{F}}$, are respectively used on $\log(S_2(l))$, $\mathcal{S}(l)$ and $\log(\mathcal{F}/3)$ of the generated field and Modane. Each discriminator uses the GAN's loss derived from the binary cross-entropy~\cite{Goodfellow2014}. We note the loss functions of these discriminators: $l_{S_2}$, $l_{\mathcal{S}}$ and $l_{\mathcal{F}}$. To define $\mathcal{D}_{S_2}$, $\mathcal{D}_{\mathcal{S}}$ and $\mathcal{D}_{\mathcal{F}}$, we use dense Neural Networks with 5 hidden layers and 25.000 parameters in total, see figure~\ref{fig:model_disc_structs}.

\begin{figure}[!htb]
\centering
\includegraphics[scale=0.8]{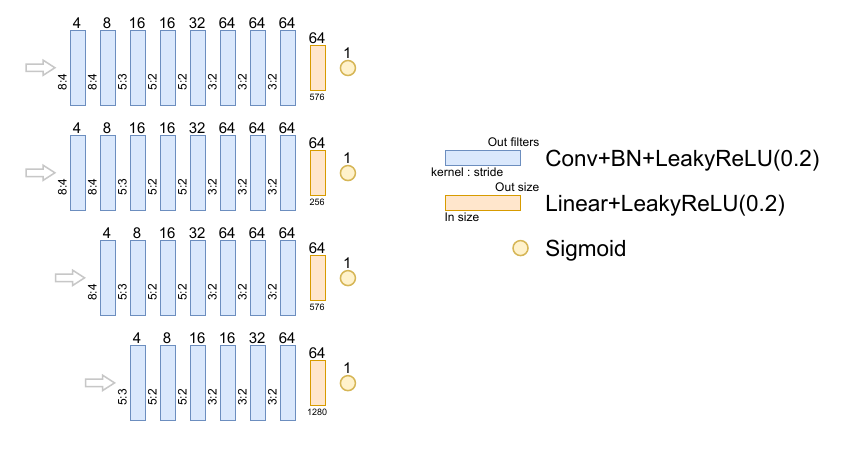}
\caption{Architecture of the four neural networks defining $\mathcal{D}_{\textrm{scale-invariance}}$. From top to bottom, each row of the discriminator takes as input a signal of size $N/2$, $N/4$, $N/8$ and $N/16$ respectively and provides a scalar value in $\left(0,1\right)$. Blue rectangles represent convolutional blocks made up of a convolutional layer, batch normalization and Leaky ReLU activation function. Yellow rectangles represent dense blocks made up of a dense layer followed by a Leaky Relu activation. The Leaky ReLu activations use a slope $0.2$ for negative values. The output size of each block is indicated by the number at the top of each rectangle. For convolutional blocks, the kernel size and stride are indicated by the numbers at the left of the blue rectangles and there is no padding. The yellow circle represents the sigmoid activation function.}
\label{fig:model_disc_scales}
\end{figure}

The proposed \textit{scale-invariance} discriminator $\mathcal{D}_{\textrm{scale-invariance}}$ is made up of four independent convolutional neural networks, each one working on portions of different length of the 1-dimensional fields, see figure~\ref{fig:model_disc_scales}. So, each NN focuses on information at different scales. Indeed, each NN works respectively on segments of length $N/2$, $N/4$, $N/8$ and $N/16$ derived from the original input of size $N$, and without overlapping between the segments of same size. Using smaller segments was found to not improve performance, likely due to the non-stationarity of the fields at smaller scales in our case study.
In this study, the smallest segments are of size $N/16 \approx L$ and so, this discriminator imposes turbulent statistics on segments of sizes going from several integral scales up to a size of the order of the integral scale.
The $\mathcal{D}_{\textrm{scale-invariance}}$ discriminator contains 197.000 parameters altogether.

The loss of $\mathcal{D}_{\textrm{scale-invariance}}$ is a weighted sum of the losses of the four convolutional networks:

\begin{equation}
    l_{SI} = \sum_{i=1}^{2} l^{(i)}_{N/2} + 0.5 \sum_{i=1}^{4} l^{(i)}_{N/4} + 0.25 \sum_{i=1}^{8} l^{(i)}_{N/8} + 0.125 \sum_{i=1}^{16} l^{(i)}_{N/16}
\end{equation}

\noindent where $l^{(i)}_{K}$ is the loss of the discriminator network applied on the ith segment of length $K$. The weights in the linear combination are added to compensate for the increasing number of segments when $K$ decreases. Each one of these losses is calculated using the GAN's loss derived from the binary cross entropy~\cite{Goodfellow2014}. 

\subsection{Optimization setup}

For training the generator and discriminators, we follow the approach proposed in~\cite{Goodfellow2014}, and consider for each discriminator its own loss function and for the generator the loss function defined as: 

\begin{equation}
    \mathcal{L} = \alpha l_{SI} + \beta l_{S_2} + \gamma l_{\mathcal{S}} + \lambda l_{\mathcal{F}}
\label{eq:loss_disc}
\end{equation}

\noindent with hyperparameters $\alpha$, $\beta$, $\gamma$ and $\lambda$. These hyper-parameters were optimized using a grid search with a step of 0.05 and three constraints: 1) the hyper-parameters must sum up to one, 2) $\alpha$ must be higher than $\beta$, $\gamma$ and $\lambda$, and 3) $\beta$ must be higher than $\gamma$ and $\lambda$. The second constraint was done to maximize the influence of the \textit{scale-invariance} discriminator over the other ones. The third constraint was added since $\mathcal{S}$ and $\mathcal{F}$ depend explicitly on $S_2$ and so, an appropriate second order structure function is needed to correctly model the skewness and flatness. The final parameters ($\alpha, \beta, \gamma, \lambda$) = (0.5, 0.2, 0.15, 0.15) were found to be optimal under the given constraints. 

GAN-Turb follows a semisupervised learning approach in which experimental turbulent data are provided only to the discriminator, but without a specific labelling. The generator only takes as input a Gaussian white noise and provides a stochastic process that has to recover some statistics of turbulence. Consequently, Modane data is not split in training, validation and test sets, and the full available dataset is used for training.

The model was trained for 500 epochs using a batch size of 32 and signals of length $N=2^{15}$. The discriminator was trained twice for each generator epoch as in~\cite{Goodfellow2014}. A learning rate of 0.001 was used for both the generator and discriminators. 
%\footnote{Training took 8.8 hours using an NVIDIA GeForce GTX 1080 (8Gb)}

\section{Results and Discussion}\label{sec:results}

In this section, we statistically characterize the stochastic field $u(x)$ provided by our model. To do so, we generate $256$ individual realizations of size $N=2^{15}$ samples. In order to avoid border effects due to the convolutional nature of our model, the noises $w(x)$ used as inputs of $\mathcal{G}$ are of size $N+N_b$, and we only keep for analysis the $N$ samples in the middle of the output field $u(x)$. $N_b$ is the number of samples impacted by border effects and in our study case $N_b=8192$. We also characterize the Modane turbulent velocity field $v(x)$ for comparison. With this purpose, we use $256$ realizations of size $N$ of the Modane turbulent dataset. 

\begin{figure}[!ht]
\centering
\includegraphics[width=\textwidth]{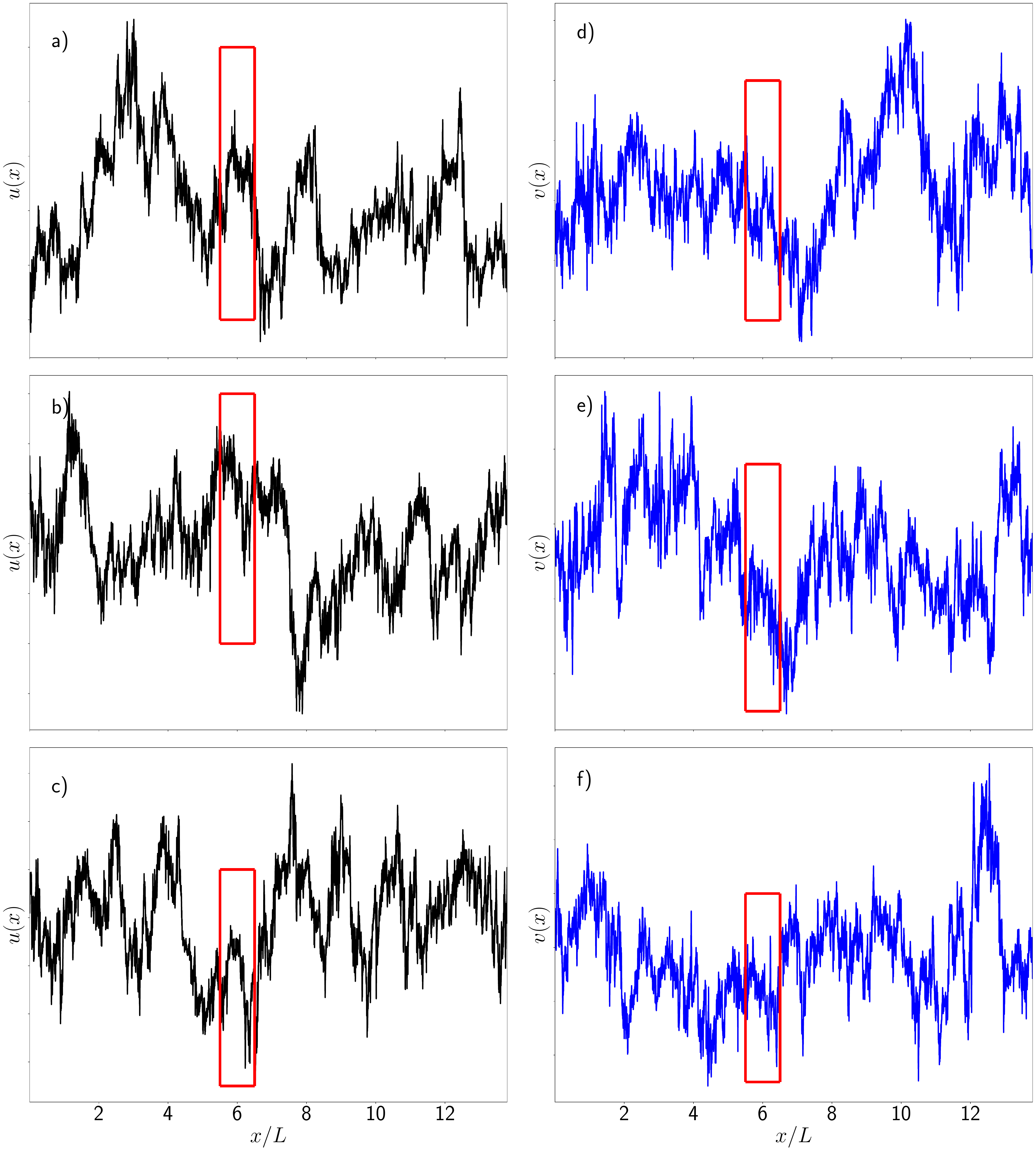}
\caption{Illustration of three realizations of the process $u(x)$ generated with our GAN approach a), b) and c) and three realizations of Modane velocity measures $v(x)$ d), e) and f), in function of the spatial variable $x/L$. The red boxes correspond to the length of a Modane integral scale $L$.}
\label{fig:Sig}
\end{figure}

Figures~\ref{fig:Sig} a), b) and c) show three realizations of $u(x)$, and d), e) and f) three realizations of $v(x)$. It is very difficult to visually distinguish the $u(x)$ and $v(x)$ fields. The only small difference comes from the slightly more intense fluctuations of Modane field at very small scale. Besides, we study the variance, skewness and flatness of the increments of the stochastic field $u(x)$ and Modane $v(x)$ across scales. We look also at the PDF of several velocity increments of $u(x)$ and $v(x)$ for scales in the integral, inertial and dissipative domains. Finally, we analyse their scaling function and compare them to the linear behavior described by the Kolmogorov 1941 theory.

\begin{figure}[!ht]
\centering
\includegraphics[width=0.55\textwidth]{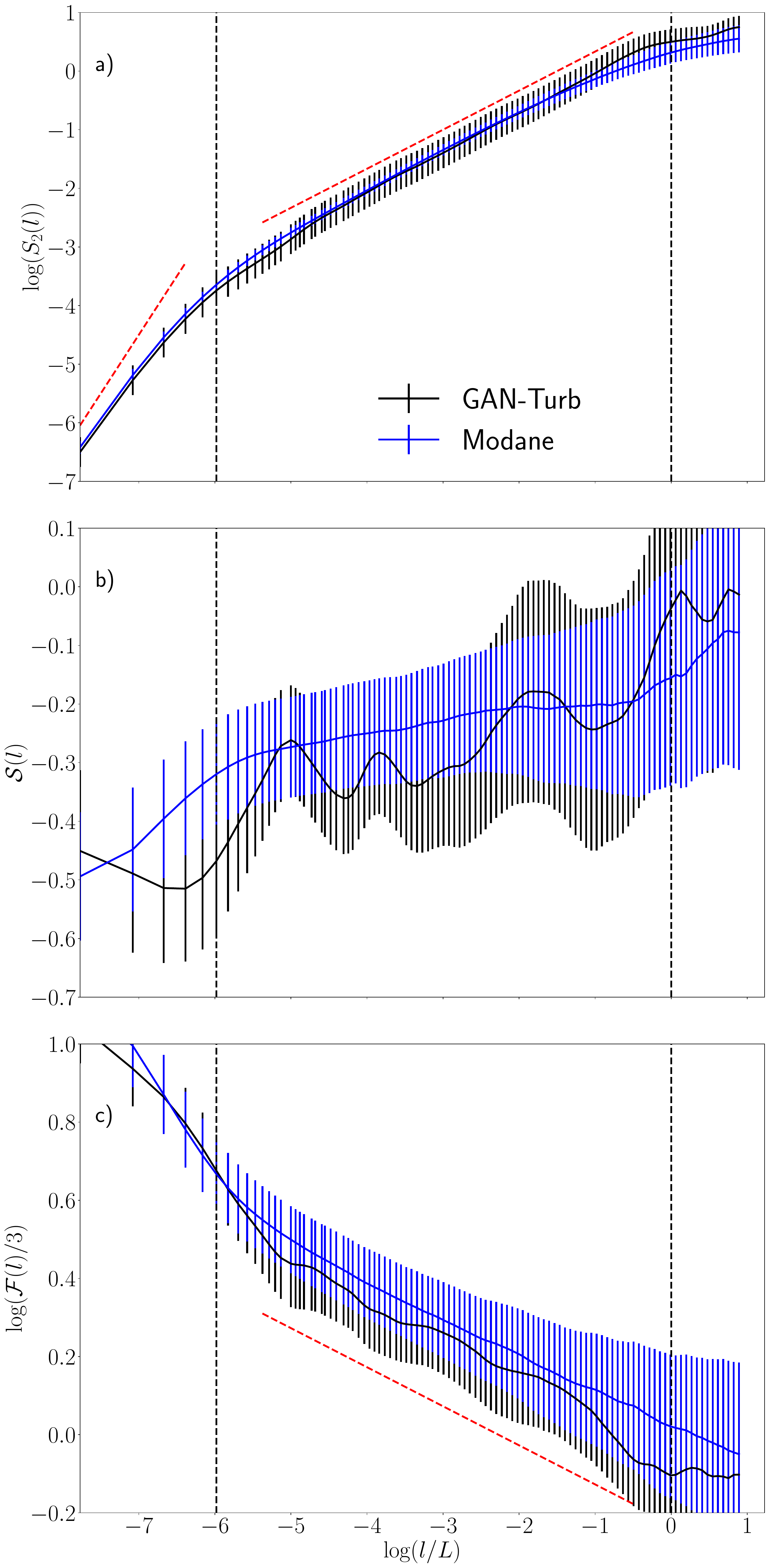}
\caption{a) Logarithm of the second order structure function $\log(S_2(l))$, b) skewness $\mathcal{S}(l)$ and c) logarithm of the flatness $\log(\mathcal{F}(l)/3)$  in function of the logarithm of the scale of analysis $\log(l/L)$ for our GAN generated field (black) and Modane (blue). Curves represent the mean value and errorbars the standard deviation calculated on 256 realizations. Red dashed lines in a) have a slope $2$ in the dissipative domain and $2/3$ in the inertial one, and describe respectively the behaviors of the Batchelor model~\cite{Batchelor1951} and the $2/3$ Kolmogorov law. Red dashed line in c) has a slope $-0.1$ previously described for the $\log(\mathcal{F}(l)/3)$ in the inertial domain~\cite{Chevillard2012}. The vertical black dashed lines correspond to the Kolmogorov $\eta$ and integral $L$ scales of Modane.}
\label{fig:Struc}
\end{figure}

Figure~\ref{fig:Struc} shows a) the logarithm of the second order structure function $\log(S_2(l))$, b) the skewness $\mathcal{S}(l)$ and c) the logarithm of the flatness $\log(\mathcal{F}(l)/3)$ in function of the logarithm of the scale of analysis $\log(l/L)$. Curves represent the mean value and errorbars the standard deviation, both calculated on 256 realizations. Blue corresponds to the Modane velocity and black to the generated field. Both the black and blue curves of $\log(S_2(l))$ are maximum at large scale $l>L$ where they present a plateau. For scales in the inertial domain $\eta<l<L$, both curves exhibit a linear behavior of slope $\sim 2/3$ up to intermittency corrections, while in the dissipative domain $l<\eta$, their slope is steeper and present a value $\sim 2$. The skewness of Modane and the generated field are close to zero at large scales $l>L$, and decrease to negative values in the inertial and dissipative domains with a steeper decrease in the dissipative domain. Finally, both curves of flatness are close to zero at large scales, illustrating the Gaussian nature of the fields in the integral domain. In the inertial domain, they increase when the scale decreases with $\log(\mathcal{F}(l)/3)$ behaving linearly with a slope $-0.1$, and this increase is steeper when entering in the dissipative domain. 

\begin{figure}[!ht]
\centering
\includegraphics[width=0.55\textwidth]{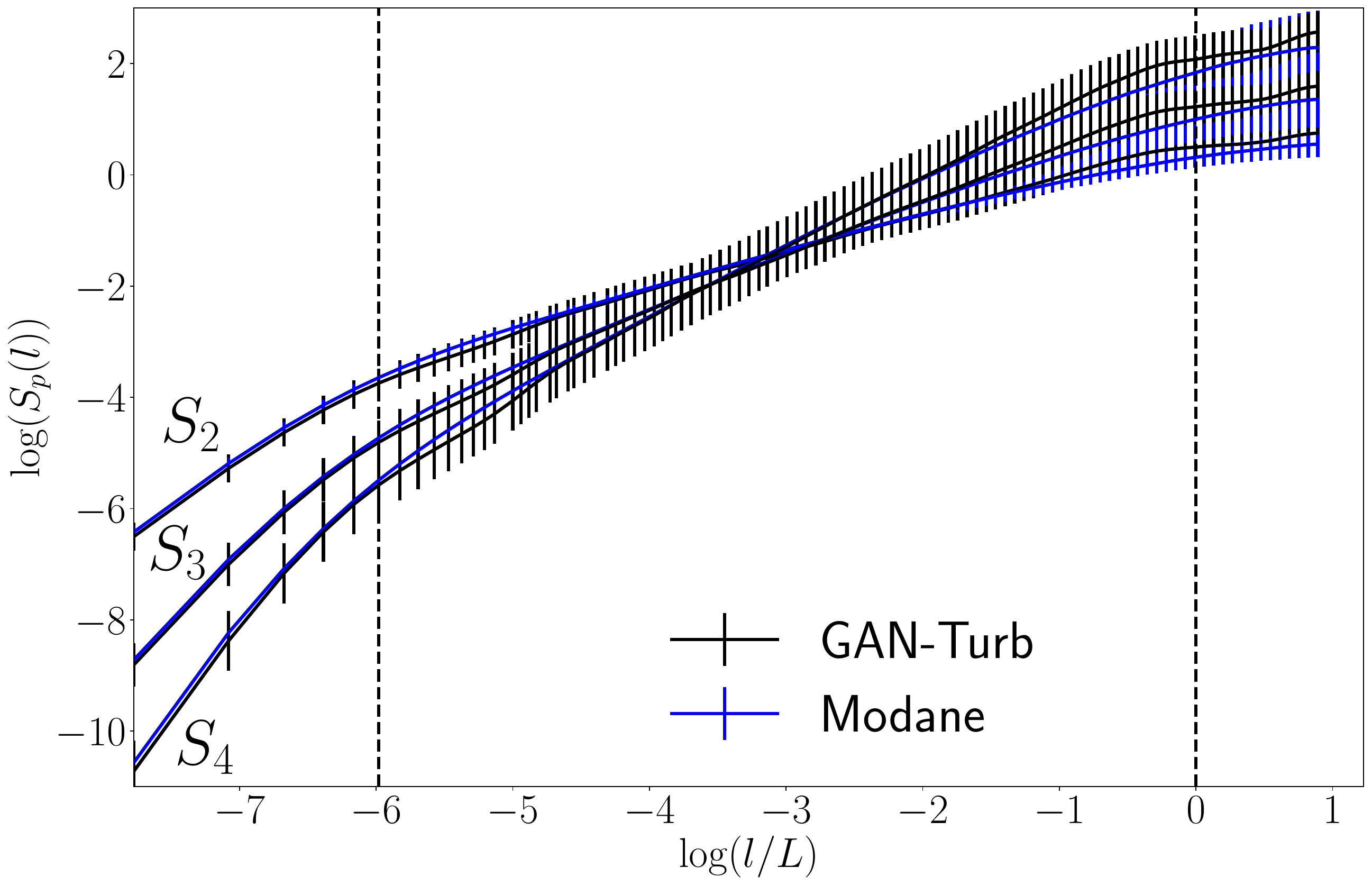}
\caption{Logarithm of the second, third and fourth order structure functions $\log(S_p(l))$  in function of the logarithm of the scale of analysis $\log(l/L)$ for our GAN generated field (black) and Modane (blue). Curves represent the mean value and errorbars the standard deviation calculated on 256 realizations. The vertical black dashed lines correspond to the Kolmogorov $\eta$ and integral $L$ scales of Modane.}
\label{fig:Struc2}
\end{figure}

So, for both Modane and the generated stochastic field, three domains of scales with different statistical behaviors are observed. For both fields, the large scales in the integral domain are Gaussian and the more energetic ones. Then, when the scale decreases the energy decreases and the PDFs become assymmetric and heavy-tailed. Furthermore, the energy distribution, skewness and flatness of $u(x)$ closely follow, within the errorbars, the ones of Modane. However, we can observe oscillations in the flatness and the skewness of the generated field that are not present in the Modane one. Actually, these measures are very sensitive to small changes in the shape of the PDFs, and point out minor deviations that are not perceptible when looking directly at the structure functions. The amplitude of the undesired oscillations is very small compared to the range of variation of the structure functions across scales, see figure~\ref{fig:Struc2}. However, when normalizing the third and fourth order structure functions by the variance at the corresponding power, the range of variation of the skewness and flatness is reduced and becomes comparable to the magnitude of the oscillations.
Estimating statistics on an increasing number of realizations of the generated stochastic field leads to a slight reduction of the amplitude of the oscillations that nevertheless still persist.
These oscillations seem to be an artifact introduced by the skewed behavior of turbulence together with the difficulty of characterizing odd statistical moments. Indeed, when studying realizations of size $N=2^{15}$ (and even larger) of Modane, these oscillations appear in the skewness. Our model seems to not be able to remove them from the generated stochastic field when increasing the number of samples on which the statistics are performed.

\begin{figure}[!htb]
\centering
\includegraphics[width=\textwidth]{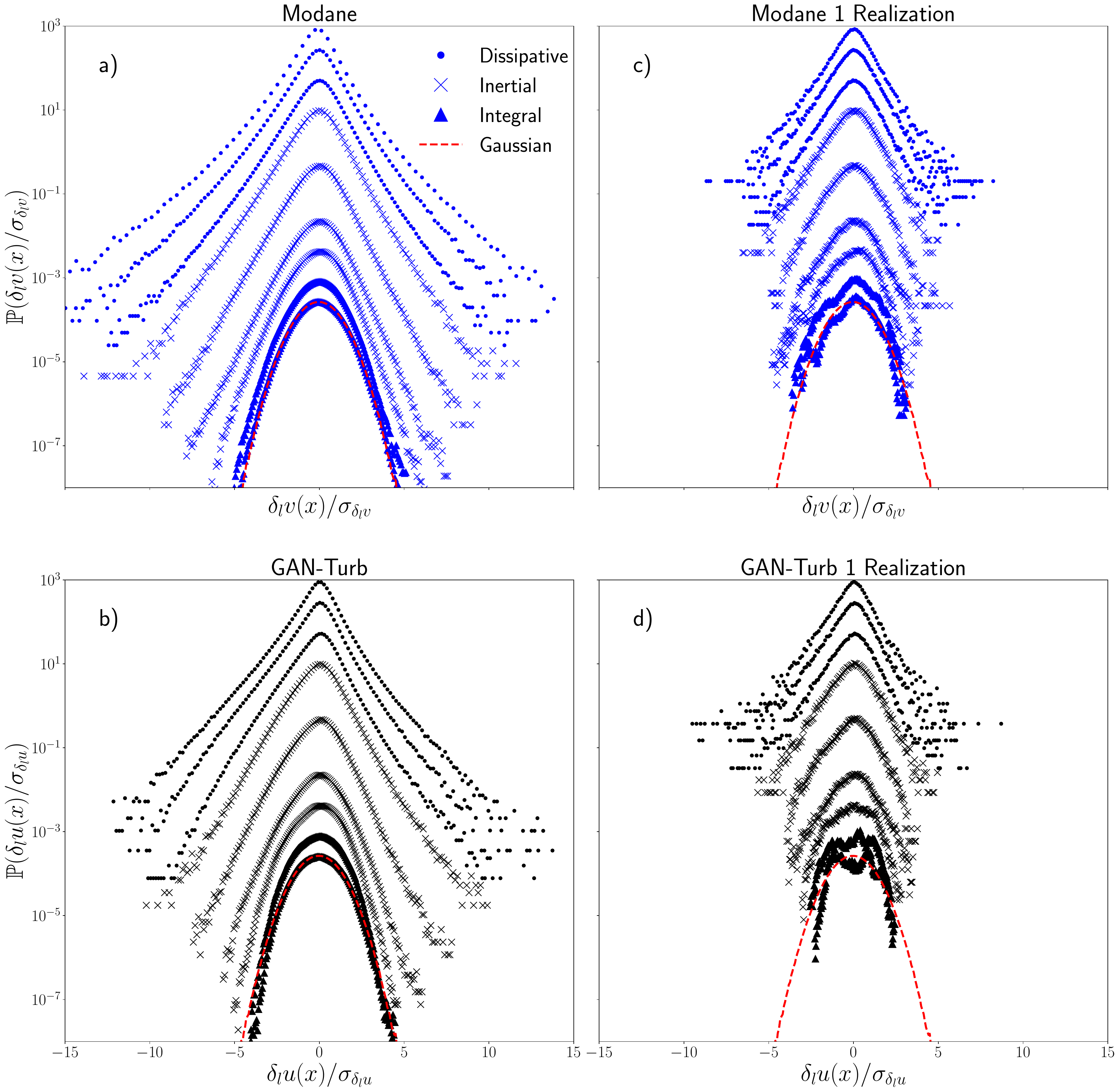}
\caption{Logarithm of the probability density function of the centered and standardized increments of the Modane turbulent velocity signal, a) and c), and the GAN fields, b) and d), in function of the values of the standardized increments. PDFs in a) and b) are obtained from 256 realizations of size $N$, and PDFs in c) and d) from only 1 realization of size $N$. The illustrated increments are those with $l = \{2,4,8,16,64,256,1024,4096,10000\} \, l_s$. The integral scale of the flow is $L=2350 \, l_s$. The red dashed line correponds to the logarithm of the probability density function of a centered and standardized Gaussian distribution.}
\label{fig:PDF}
\end{figure}

These results are supported by figure~\ref{fig:PDF}, which illustrates the PDF of several standardized velocity increments of Modane a) and c) and the generated field b) and d) for scales in the integral, inertial and dissipative domains. While the histograms of a) and b) are done taking into account respectively the whole Modane and generated dataset (256 realizations of size $N$), the histograms of c) and d) are done with only one segment of size $N$, as the ones used during training. We compare first the full PDFs of Modane a) and the generated field b). Both processes present close to Gaussian PDFs at scales in the integral domain which become the more and more assymmetric and heavy-tailed when the scale decreases through the inertial and dissipative domains. The deformation of the PDF of $u(x)$ across scales is very close to the one of Modane. However, in the integral domain, the PDF of $u(x)$ has slightly fewer tails than a Gaussian, and in the dissipative domain, the Modane PDFs present stronger extreme events than the ones of the generated field. Indeed, our generative model is trained with single realizations of size $N$ of Modane, which exhibit few extreme events at small scales as shown by figure~\ref{fig:PDF} c). Looking at single realizations, the PDFs of the generated field d) are almost indistinguishable from the Modane ones c), and the few extreme events are well captured. This explains the good similarity between PDFs in a) and b). Nevertheless, still rarer extreme events in Modane are not learned by our model and lead to the slightly heavier tails of a) compared to b). To complete this visual inspection, we estimated the Jeffreys distance $\mathcal{J}$ between the PDFs of the standardized velocity increments of Modane and the generated process, with sizes ranging from the dissipative to the integral domains. We considered the PDFs built with $256$ realizations and the Jeffreys distance as defined in~\cite{Chung1989}. By construction, this distance in bounded between $0$ (for identical PDFs) and $2$, and we obtained negligeable values for all the studied increments, see~\ref{appendix}.

\begin{figure}[!htb]
\centering
\includegraphics[width=0.6\textwidth]{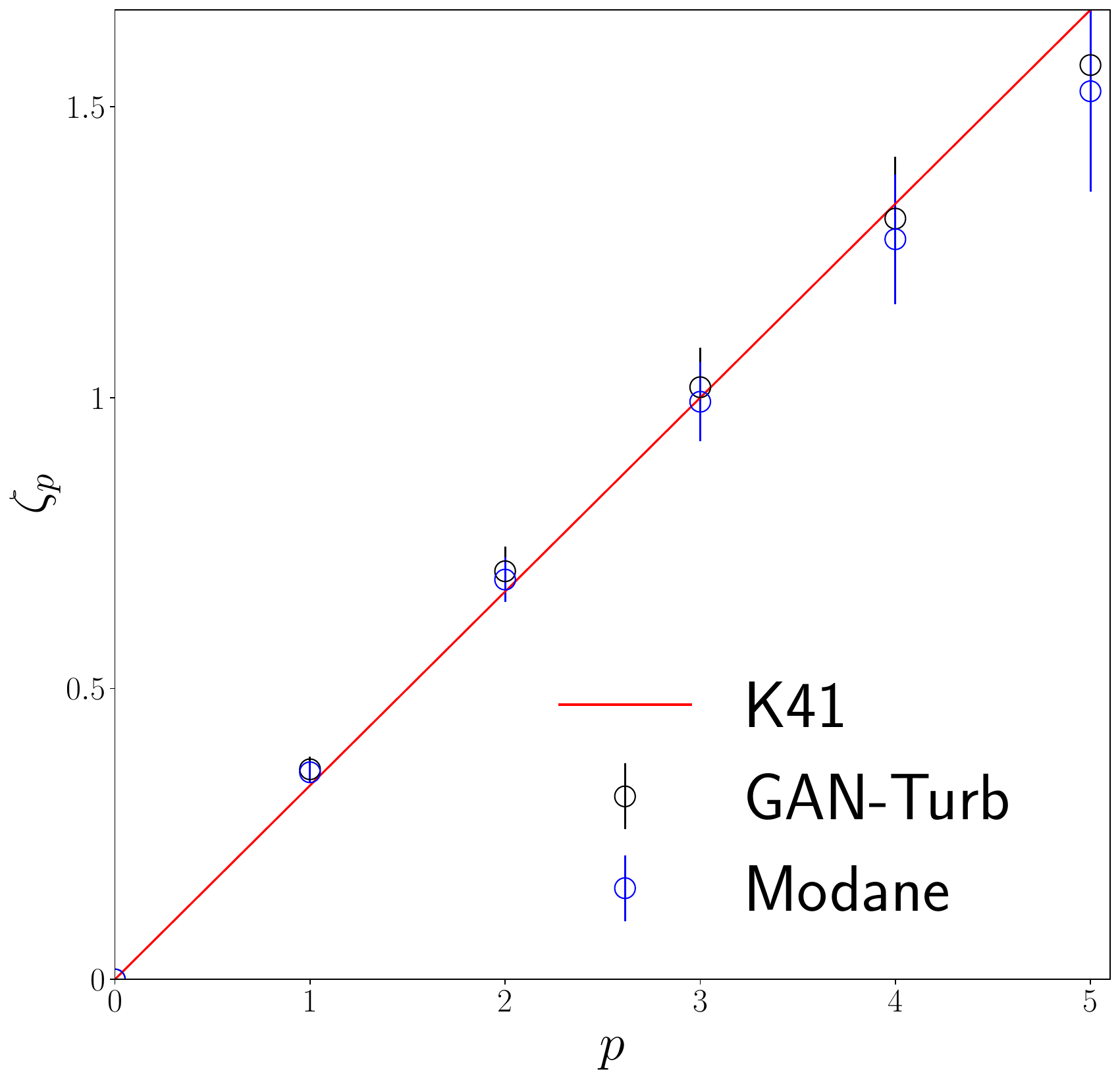}
\caption{Scaling function, $\zeta_p$, in function of $p$ for the Kolmogorov 1941 model (red), Modane velocity turbulent dataset (blue) and the GAN-Turb velocity field (black).}
\label{fig:scaling}
\end{figure}

Figure~\ref{fig:scaling} shows the scaling exponent $\zeta_p$ in function of $p$ for Modane in blue and the generated field in black, as well as the behavior predicted by the Kolmogorov 1941 theory in red. The scaling exponents of $v(x)$ and $u(x)$ are obtained by fitting the slope of $\log(S_p(l))$ vs $\log(l)$ in the inertial region far away from $\eta$ and $L$. In this study, we used $17 l_s<l<274 l_s$ to define the fitting region. Both Modane and $u(x)$ present a non-linear scaling function matching within the errorbars. Finally, $u(x)$ has $\zeta_3=1$ and so, it respects the $4/5$ law of Kolmogorov.

Two alternative models from the state of the art were also tested. First, a classical GAN model~\cite{Goodfellow2014} and second a Wasserstein GAN (WGAN) model~\cite{Arjovsky2017}, both with the generator presented in section~\ref{sec:generator} and the discriminator following the architecture of one of the networks of the \textit{scale-invariance} discriminator of section~\ref{sec:discriminators} but applied directly on signals of size $N$. Results presented in~\ref{appendix} show that these models do not allow to generate stochastic fields with turbulent velocity statistics. This illustrates the contribution of our multiscale multicriteria physics based approach.

\section{Conclusions}\label{sec:conclusions}

We present a neural network generator model embedded in a multicriteria GAN learning strategy for the synthesis of a 1-dimensional stochastic field with turbulent statistics. 
The presented approach is guided by the multiscale physics of turbulence in two ways. First, the generator architecture is fully convolutional, therefore it mimicks the multiresolution analysis usually done to characterize turbulent velocity across scales. Second, the training schema is based on four discriminators, each one focusing on a given statistical property of turbulence: 1) variance, 2) skewness and 3) flatness of the velocity increments across scales, and 4) the full PDF of the velocity field at different scales of resolution. The three first discriminators are respectively linked to the energy distribution, energy cascade and intermittency phenomenon. 

Consequently, this model is able to generate an intermittent and scale-invariant stochastic field $u(x)$. This field exhibits three domains of scales with different statistical behavior as well as the energy distribution and the energy cascade of turbulence as illustrated by $S_2(l)$ and $\mathcal{S}(l)$ respectively. Furthermore, the PDF of the increments of the field deformates from Gaussian at large scale to heavy-tailed and asymmetric at small scales, such as the PDF of the turbulent velocity increments. This is illustrated by $\mathcal{S}(l)$, $\mathcal{F}(l)$ and $\zeta_p$.
Moreover, looking at the PDFs of the increments of $u(x)$ computed on $256$ realizations of size $N$, we observe that the model is able to recover extreme events that do not appear in single realizations of the same size, as the ones used during training. 
Contrary to most of previous work on neural network modelling of turbulence~\cite{Wu2020,Yousif2021,Yousif2022,Subel2023}, we not only focus on second order statistics of the velocity field, but also on higher-order statistics. More precisely, we study the skewness and flatness of the velocity increments that are theoretically linked with the energy cascade and intermittency.

We compare the stochastic field generated with our physics based multiscale approach to two stochastic fields generated by state of the art methods: a classical GAN and a Wasserstein GAN. Our approach clearly outperforms the state of the art ones.

Stochastic generative models, such as the one presented in this work, allow the generation of large amounts of data with specific statistics, which are useful for validating methodologies of analysis~\cite{GraneroBelinchon2019a} or feeding machine learning algorithms when real data are scarce~\cite{Chen2021,Gao2023}.
Furthermore, generative models provide new insights into the studied process itself by characterizing how the model operates to produce the stochastic field~\cite{Subel2023}.
In addition, many processes from different fields such as geophysics~\cite{Khouider2003,Chapron2018}, finance~\cite{Muzy2000,Bacry2001,Han2016}, biomedicine~\cite{OchabMarcinek2004,Baar2016}, biology~\cite{Schmitt2001} or telecommunications~\cite{Gloaguen2006,Eisenblatter2012} present complex dynamics requiring higher-order statistics for a correct modelling, and so, our model can find applications in a large range of domains.

Finally, the main future perspective is to generalize the proposed approach to synthesize 2d stochastic fields with the statistics of homogeneous and isotropic turbulent velocity. 
Furthermore, we consider to generalize the proposed generative model by training it on different datasets at different Reynolds numbers and by conditioning the model by the Reynolds number of the flow.
The multiscale and multicriteria physics based GAN used in this work is available at https://github.com/manuelcgallucci/DCGAN-turb.

\ack
This work was supported by the French National Research Agency (ANR-21-CE46-0011-01), within the program ”Appel \`a projets g\'en\'erique 2021”.

\appendix

\section{Results with GAN and Wasserstein GAN models}\label{appendix}

This section illustrates the performances of two neural network models from the state of the art, a classical GAN and a Wasserstein GAN, when dealing with the generation of a 1-dimensional stochastic field with turbulent statistics. Both models use the generator presented in section~\ref{sec:generator} and a discriminator with the architecture of one of the networks of the \textit{scale-invariance} discriminator of section~\ref{sec:discriminators} but applied directly on the generated and Modane signals. 

\begin{figure}[!htb]
\centering
\includegraphics[width=\textwidth]{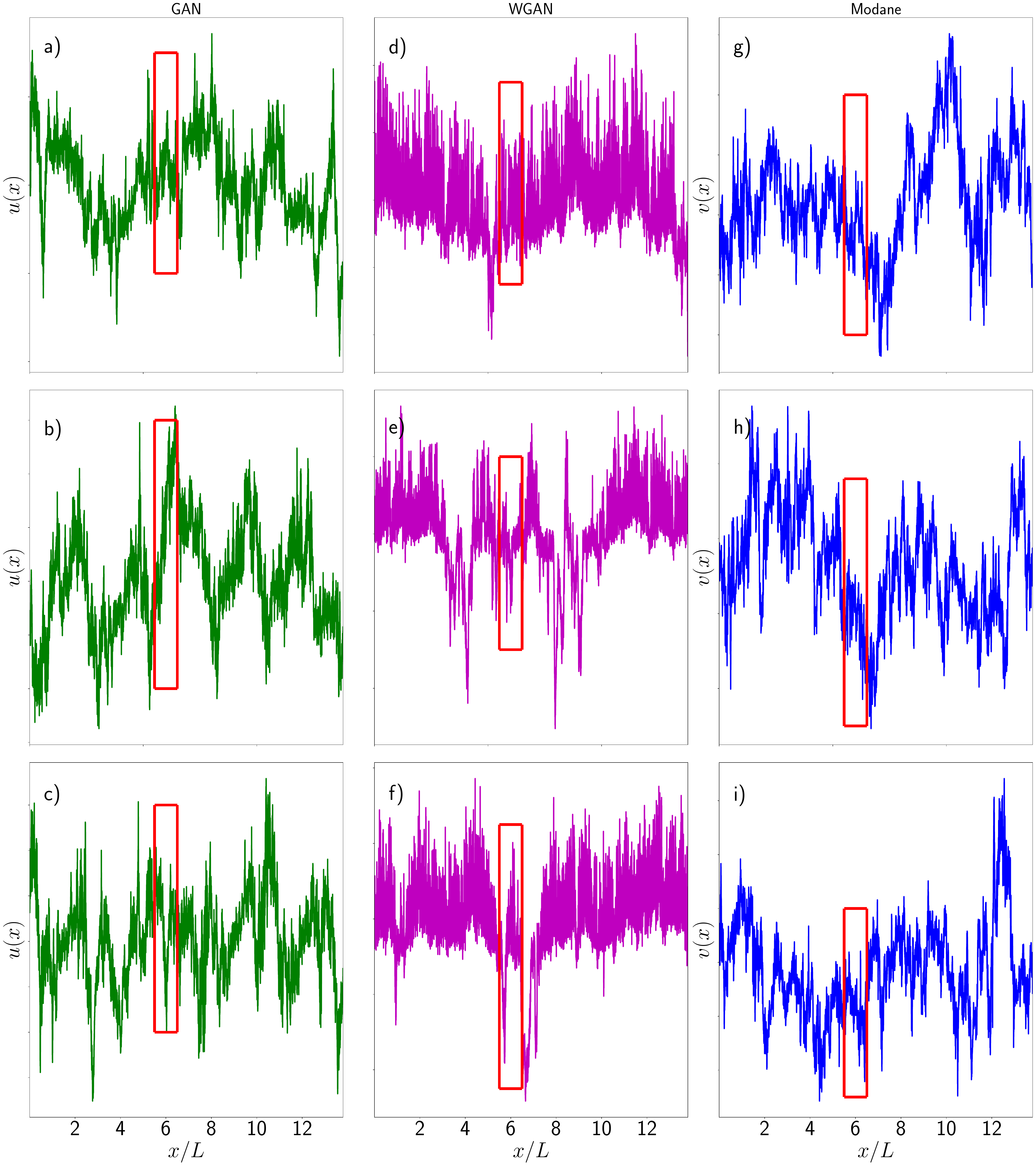}
\caption{Illustration of three realizations of the process $u(x)$ generated with a classical GAN a), b) and c) and a Wasserstein GAN d), e) and f), and three realizations of Modane velocity field g), h) and i), in function of the spatial variable $x/L$. The red boxes correspond to the length of a Modane integral scale $L$.}
\label{fig:SigBasic}
\end{figure}

Figure~\ref{fig:SigBasic} shows three realizations of the GAN and WGAN models in green and magenta respectively, and three realizations of the Modane turbulent velocity field in blue for comparison. While the GAN field is visually very close to Modane, with large and small scale structures of similar sizes, this is not the case for the WGAN, which presents very energetic small scales and few large scale structures.

\begin{figure}[!htb]
\centering
\includegraphics[width=0.55\textwidth]{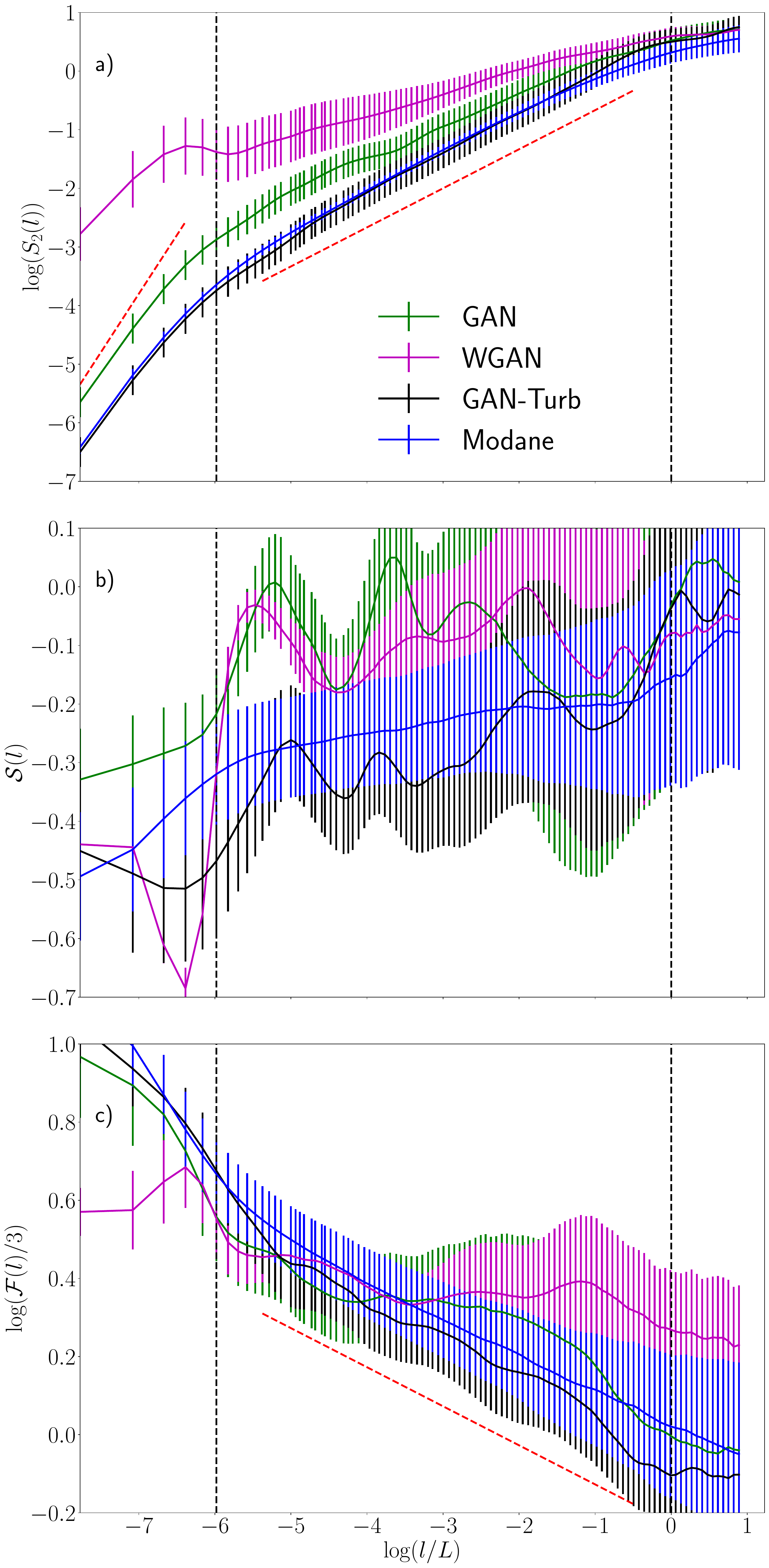}
\caption{a) Logarithm of the second order structure function $\log(S_2(l))$, b) skewness $\mathcal{S}(l)$ and c) logarithm of the flatness $\log(\mathcal{F}(l)/3)$  in function of the logarithm of the scale of analysis $\log(l/L)$ for a stochastic field generated with a classical GAN (green), a Wasserstein GAN (magenta), GAN-Turb (black) and for the Modane velocity field (blue). Curves represent the mean value and errorbars the standard deviation calculated on 256 realizations. Red dashed lines in a) have a slope $2$ in the dissipative domain and $2/3$ in the inertial one, that describe respectively the behaviors of the Batchelor model~\cite{Batchelor1951} and the $2/3$ Kolmogorov law. Red dashed line in c) has a slope $-0.1$ previously described for the $\log(\mathcal{F}(l)/3)$ in the inertial domain~\cite{Chevillard2012}. The vertical black dashed lines correspond to the Kolmogorov $\eta$ and integral $L$ scales of Modane.}
\label{fig:StrucBasic}
\end{figure}

This visual conclusion is supported by figure~\ref{fig:StrucBasic} a) which shows $\log(S_2(l))$ in function of the logarithm of the scale of analysis for the GAN (green), WGAN (magenta), GAN-Turb (black) and Modane (blue) fields. Both the GAN and WGAN fields present small scales that are too energetic compared to Modane, and slopes in the inertial domain smaller than the expected $2/3$ value. Figures~\ref{fig:StrucBasic} b) and c) present respectively the skewness and flatness of the GAN, WGAN, GAN-Turb and Modane fields. For both GAN and WGAN the skewness and flatness strongly oscillate, and are far from the Modane ones in several domains of scales.

\begin{figure}[!htb]
\centering
\includegraphics[width=\textwidth]{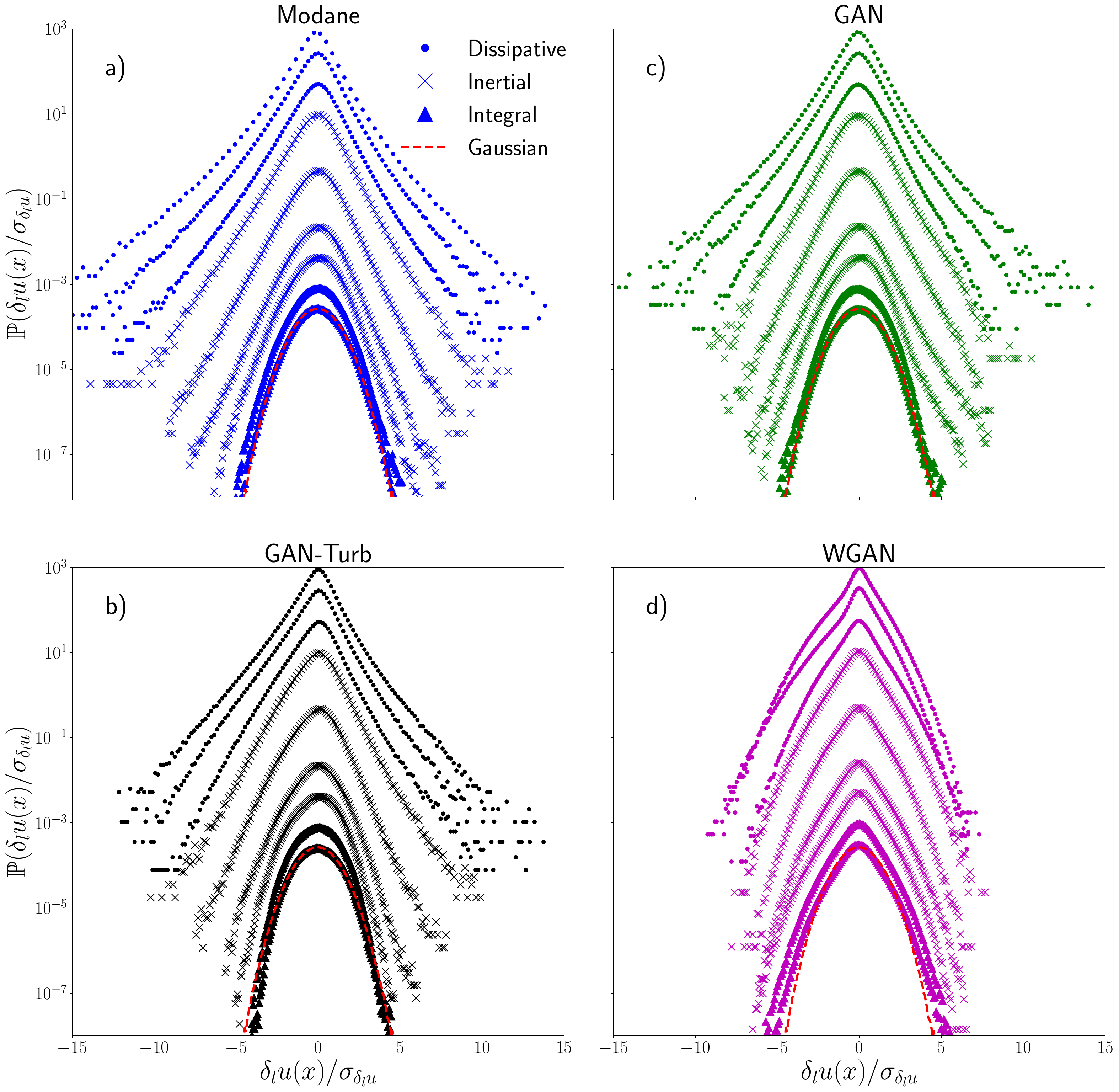}
\caption{Logarithm of the probability density function of the centered and standardized increments of the Modane turbulent velocity signal a), GAN-Turb b), a classical GAN field c) and a Wasserstein GAN field d), in function of the values of the standardized increments. The PDFs are obtained from 256 realizations of size $N$ of the fields. The illustrated increments are those with $l = \{2,4,8,16,64,256,1024,4096,10000\} \, l_s$. The integral scale of the flow is $L=2350 \, l_s$. The red dashed line correponds to the logarithm of the probability density function of a centered and standardized Gaussian distribution.}
\label{fig:PDFBasic}
\end{figure}

\begin{figure}[!htb]
\centering
\includegraphics[width=0.55\textwidth]{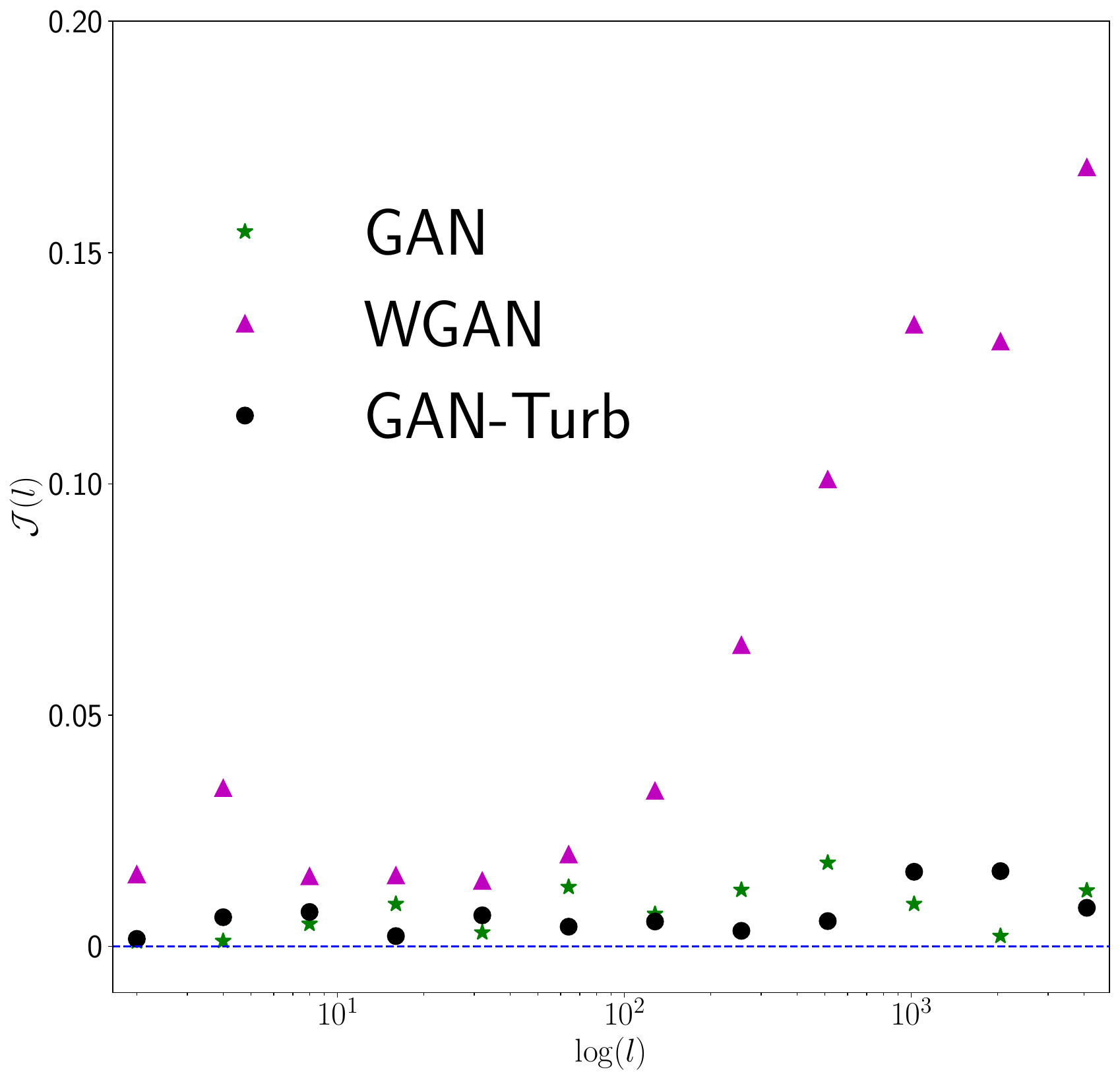}
\caption{Jeffreys distance, as defined in~\cite{Chung1989}, between the probability density function of the centered and standardized increments of the Modane turbulent velocity signal and GAN-Turb (black), a classical GAN field (green) and a Wasserstein GAN field (magenta). The Jeffreys distance is estimated on the PDFs obtained from 256 realizations of size $N$ of the fields. The sizes of the increments are $l = \{2,4,8,16,32,64,128,256,512,1024,2048,4096\} \, l_s$. The integral scale of the flow is $L=2350 \, l_s$. The blue dashed horizontal line correponds to $0$.}
\label{fig:Hell}
\end{figure}

Figure~\ref{fig:PDFBasic} presents the logarithm of the PDF of the standardized and centered increments of the GAN, WGAN, GAN-Turb and Modane velocity fields. On the one hand, the stochastic field generated with the WGAN model presents PDFs with shapes that are very different from Modane. On the other hand, the shapes of the PDFs of the field generated with the classical GAN are visually very similar to Modane. However, even for the GAN field, figure~\ref{fig:StrucBasic} quantifies the strong differences between the PDFs of Modane and the classical GAN field. This is specially true for the evolution of the variances of the increments, illustrated by $\log(S_2(l))$, which are not visible in~\ref{fig:PDFBasic}. To complete the study of PDFs, figure~\ref{fig:Hell} shows the Jeffreys distance $\mathcal{J}$, as defined in~\cite{Chung1989}, between the PDFs of several standardized increments of Modane and GAN (green), WGAN (magenta) and GAN-Turb (black) fields. The size of the increments ranges from the dissipative to the integral domain.

\begin{figure}[!htb]
\centering
\includegraphics[width=0.55\textwidth]{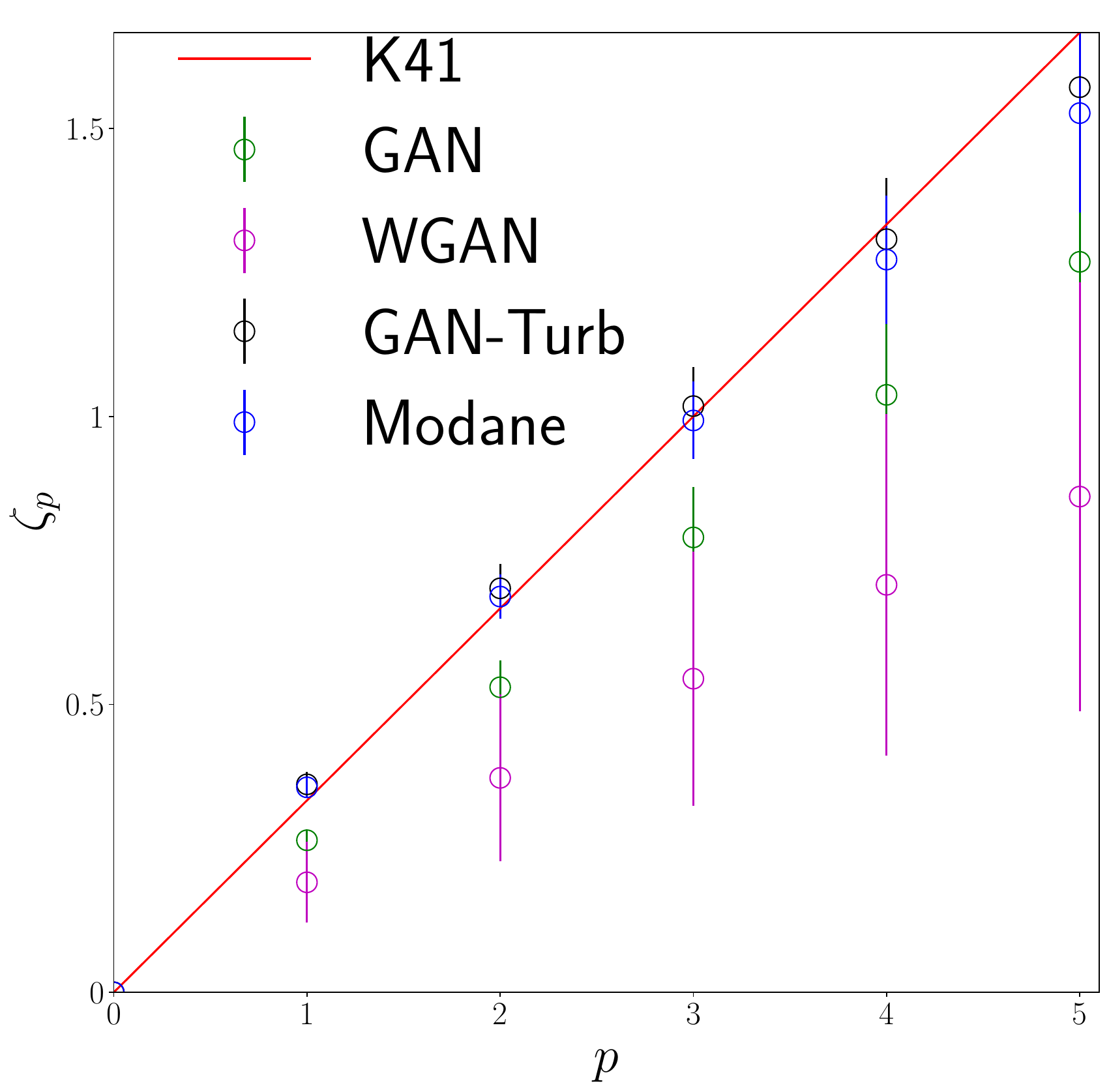}
\caption{Scaling function, $\zeta_p$, in function of $p$ for the Kolmogorov 1941 model (red), Modane velocity turbulent dataset (blue), GAN-Turb (black), classical GAN velocity field (green) and Wasserstein GAN field (magenta).}
\label{fig:scalingBasic}
\end{figure}

Figure~\ref{fig:scalingBasic} shows the behavior of the scaling function $\zeta_p$ in function of $p$ for the Modane (blue), GAN (green), WGAN (magenta) and GAN-Turb (black) field, and provides a clear illustration of the absence of physical sens of the fields generated with GAN and WGAN. As said in section~\ref{sec:turb}, $\zeta_3$ must be equal to $1$ for turbulence, and this is not the case for either the classical GAN or the WGAN.

So, we conclude that our GAN-Turb model outperforms state of the art ones, and generates a physical stochastic field with turbulent statistics. 

\clearpage

\bibliographystyle{unsrt} 

\bibliography{THEBIBLIO}

\begin{thebibliography}{10}

\bibitem{Kolmogorov1991}
A.~N. Kolmogorov.
\newblock The local structure of turbulence in incompressible viscous fluid for
  very large {R}eynolds numbers.
\newblock {\em Proceedings: Mathematical and Physical Sciences},
  434(1890):9--13, 1991.

\bibitem{Kolmogorov1962}
A.~N. Kolmogorov.
\newblock A refinement of previous hypotheses concerning the local structure of
  turbulence in a viscous incompressible fluid at high {R}eynolds number.
\newblock {\em Journal of Fluid Mechanics}, 13:82--85, 1962.

\bibitem{Obukhov1962}
A.~M. Obukhov.
\newblock Some specific features of atmospheric turbulence.
\newblock {\em Journal of Fluid Mechanics}, 13:77--81, 1962.

\bibitem{Frisch1995}
U.~Frisch.
\newblock {\em Turbulence: the legacy of {A.N. Kolmogorov}}.
\newblock Cambridge {U}niversity {P}ress, 1995.

\bibitem{Atta1980}
C.~W.~Van Atta and R.~A. Antonia.
\newblock {Reynolds number dependence of skewness and flatness factors of
  turbulent velocity derivatives}.
\newblock {\em The Physics of Fluids}, 23(2):252--257, 1980.

\bibitem{Anselmet1984}
F.~Anselmet, Y.~Gagne, E.~J. Hopfinger, and R.~A. Antonia.
\newblock High-order velocity structure functions in turbulent shear flows.
\newblock {\em Journal of Fluid Mechanics}, 140:63--89, 1984.

\bibitem{Gagne1990}
Y.~Gagne, E.~J. Hopfinger, and U.~Frisch.
\newblock {\em New trends in nonlinear dynamics and pattern-forming phenomena},
  volume 237 of {\em NATO ASI Series}, chapter A new universal scaling for
  fully developed turbulence: the distribution of velocity increments, pages
  315--319.
\newblock Springer, New York, 1990.

\bibitem{Meneveau1991}
C.~Meneveau and K.R. Sreenivasan.
\newblock The multifractal nature of turbulent energy dissipation.
\newblock {\em Journal of Fluid Mechanics}, 224:429--484, 1991.

\bibitem{Lohse1993}
D.~Lohse and S.~Grossmann.
\newblock Intermittency in turbulence.
\newblock {\em Physica A: Statistical Mechanics and its Applications},
  194(1):519--531, 1993.

\bibitem{Flandrin1989}
P.~Flandrin.
\newblock On the spectrum of fractional {B}rownian motions.
\newblock {\em IEEE Transactions on Information Theory}, 35(1):197--199, 1989.

\bibitem{Flandrin1992}
P.~Flandrin.
\newblock Wavelet analysis and synthesis of fractional {B}rownian motion.
\newblock {\em IEEE Transactions on Information Theory}, 38(2):910--917, 1992.

\bibitem{Benzi1993}
R.~Benzi, L.~Biferale, A.~Crisanti, G.~Paladin, M.~Vergassola, and A.~Vulpiani.
\newblock A random process for the construction of multiaffine fields.
\newblock {\em Physica D}, 65(4):352--358, 1993.

\bibitem{Arneodo1998}
A.~Arneodo, E.~Bacry, and J.~F. Muzy.
\newblock Random cascades on wavelet dyadic trees.
\newblock {\em Journal of Mathematical Physics}, 39(8):4142--4164, 1998.

\bibitem{Bacry2001}
E.~Bacry, J.~Delour, and J.~F. Muzy.
\newblock Multifractal random walk.
\newblock {\em Physical Review E}, 64:026103, 2001.

\bibitem{Robert2008}
R.~Robert and V.~Vargas.
\newblock Hydrodynamic turbulence and inter-mittent random fields.
\newblock {\em Communications in Mathematical Physics}, 284:649, 2008.

\bibitem{Chevillard2012}
L.~Chevillard, B.~Castaing, A.~Arneodo, E.~L{\'e}v{\^e}que, J.F. Pinton, and
  S.G. Roux.
\newblock A phenomenological theory of eulerian and lagrangian velocity
  fluctuations in turbulent flows.
\newblock {\em Comptes Rendus Physique}, 13(9):899 -- 928, 2012.

\bibitem{Du2018}
Y.~Du and G.~Lin.
\newblock Turbulence generation from a stochastic wavelet model.
\newblock {\em Proceedings of the Royal Society A}, 474:20180093, 2018.

\bibitem{Chevillard2019}
L.~Chevillard, C.~Garban, R.~Rhodes, and V.~Vargas.
\newblock On a skewed and multifractal unidimensional random field, as a
  probabilistic representation of {K}olmogorov's views on turbulence.
\newblock {\em Annales Henri Poincar{\'e}}, 20:3693--3741, 2019.

\bibitem{Peinke2019}
J.~Peinke, M.~R.~R. Tabar, and M.~Wachter.
\newblock The {F}okker–{P}lanck approach to complex spatiotemporal disordered
  systems.
\newblock {\em Annual Review of Condensed Matter Physics}, 10(1):107--132,
  2019.

\bibitem{Alexandrov2020}
A.~V. Alexandrov, L.~W. Dorodnicyn, A.~P. Duben, and D.~R. Kolyukhin.
\newblock Generation of the stochastic anisotropic velocity field for turbulent
  flow simulation.
\newblock In {\em AIP Conference Proceedings}, volume 2312, page 050001, 2020.

\bibitem{Mandelbrot1968}
B.B. Mandelbrot and J.W. Van~Ness.
\newblock Fractional brownian motions fractional noises and applications.
\newblock {\em SIAM Review}, 10(4):422--437, 1968.

\bibitem{Peinke1993}
J.~Peinke, M.~Klein, A.~Kitte, A.~Okninsky, J.~Parisi, and 0.~E. Roessler.
\newblock On chaos, fractals and turbulence.
\newblock {\em Physica Scripta}, T49:672--676, 1993.

\bibitem{Nawroth2006}
A.~P. Nawroth and J.~Peinke.
\newblock Multiscale reconstruction of time series.
\newblock {\em Physics Letters A}, 360(2):234--237, 2006.

\bibitem{Klein2003}
M.~Klein, A.~Sadiki, and J.~Janicka.
\newblock A digital filter based generation of inflow data for spatially
  developing direct numerical or large eddy simulations.
\newblock {\em Journal of Computational Physics}, 186(2):652--665, 2003.

\bibitem{Hoepffner2011}
J.~Hoepffner, Y.~Naka, and K.~Fukagata.
\newblock Realizing turbulent statistics.
\newblock {\em Journal of Fluid Mechanics}, 676:54--80, 2011.

\bibitem{Mann1998}
Jakob Mann.
\newblock Wind field simulation.
\newblock {\em Probabilistic Engineering Mechanics}, 13(4):269--282, 1998.

\bibitem{Goodfellow2014}
I.~Goodfellow, J.~Pouget-Abadie, M.~Mirza, B.~Xu, D.~Warde-Farley, S.~Ozair,
  A.~Courville, and Y.~Bengio.
\newblock Generative {A}dversarial {N}ets.
\newblock In {\em Advances in Neural Information Processing Systems},
  volume~27, 2014.

\bibitem{Beroud2023}
T.~Beroud, P.~Abry, Y.~Malevergne, M.~Senneret, G.~Perrin, and J.~Macq.
\newblock Wassertein {GAN} synthesis for time series with complex temporal
  dynamics: frugal architectures and arbitrary sample-size generation.
\newblock In {\em IEEE International Conference on Acoustics, Speech and Signal
  Processing (ICASSP)}, pages 1--5, Rhodes Island, Greece, 2023.

\bibitem{Song2019}
Y.~Song and S.~Ermon.
\newblock Generative modeling by estimating gradients of the data distribution.
\newblock In {\em Advances in Neural Information Processing Systems},
  volume~32, 2019.

\bibitem{Yan2021}
T.~Yan, H.~Zhang, T.~Zhou, Y.~Zhan, and Y.~Xia.
\newblock Score{G}rad: multivariate probabilistic time series forecasting with
  continuous energy-based generative models, 2021.

\bibitem{Dhariwal2021}
P.~Dhariwal and A.~Nichol.
\newblock Diffusion models beat {GAN}s on image synthesis.
\newblock In {\em Advances in Neural Information Processing Systems},
  volume~34, pages 8780--8794, 2021.

\bibitem{Zhang2020}
Z.~Zhang, R.~Zhang, Z.~Li, Y.~Bengio, and L.~Paull.
\newblock Perceptual generative autoencoders.
\newblock In {\em Proceedings of the 37th International Conference on Machine
  Learning,}, volume 119, pages 11298--11306, 2020.

\bibitem{Ye2022}
F.~Ye and A.~G. Bors.
\newblock Deep mixture generative autoencoders.
\newblock {\em IEEE Transactions on Neural Networks and Learning Systems},
  33(10):5789--5803, 2022.

\bibitem{Brunton2020}
S.~L. Brunton, B.~R. Noack, and P.~Koumoutsakos.
\newblock Machine learning for fluid mechanics.
\newblock {\em Annual Review of Fluid Mechanics}, 52:477--508, 2020.

\bibitem{Beck2021}
A.~Beck and M.~Kurz.
\newblock A perspective on machine learning methods in turbulence modeling.
\newblock {\em Surveys for Applied Mathematics and Mechanics}, 44:e202100002,
  2021.

\bibitem{Li2023a}
T.~Li, M.~Buzzicotti, L.~Biferale, and F.~Bonaccorso.
\newblock Generative adversarial networks to infer velocity components in
  rotating turbulent flows.
\newblock {\em The European Physical Journal E}, 31:46, 2023.

\bibitem{Li2023b}
T.~Li, L.~Biferale, F.~Bonaccorso, M.~A. Scarpolini, and M.~Buzzicotti.
\newblock Synthetic lagrangian turbulence by generative diffusion models, 2023.

\bibitem{Deng2019}
Z.~Deng, C.~He, Y.~Liu, and K.~C. Kim.
\newblock Super-resolution reconstruction of turbulent velocity fields using a
  generative adversarial network-based artificial intelligence framework.
\newblock {\em Physics of Fluids}, 31:125111, 2019.

\bibitem{Liu2020}
B.~Liu, J.~Tang, H.~Huang, and X.~Y. Lu.
\newblock Deep learning methods for super-resolution reconstruction of
  turbulent flows.
\newblock {\em Physics of fluids}, 32(2):025105, 2020.

\bibitem{Kim2020}
J.~Kim and C.~Lee.
\newblock Deep unsupervised learning of turbulence for inflow generation at
  various {R}eynolds numbers.
\newblock {\em Journal of Computational Physics}, 406:109216, 2020.

\bibitem{Kim2021}
H.~Kim, J.~Kim, S.~Won, and C.~Lee.
\newblock Unsupervised deep learning for super-resolution reconstruction of
  turbulence.
\newblock {\em Journal of Fluid Mechanics}, 910:A29, 2021.

\bibitem{Geneva2020}
N.~Geneva and N.~Zabaras.
\newblock Multi-fidelity generative deep learning turbulent flows.
\newblock {\em Foundations of Data Science}, 2:391--428, 2020.

\bibitem{Drygala2022}
C.~Drygala, B.~Winhart, F.~di~Mare, and H.~Gottschalk.
\newblock Generative modeling of turbulence.
\newblock {\em Physics of Fluids}, 34:035114, 2022.

\bibitem{Wang2020}
R.~Wang, K.~Kashinath, M.~Mustafa, A.~Albert, and R.~Yu.
\newblock Towards physics-informed deep learning forturbulent flow prediction.
\newblock In {\em In {P}roceedings of the 26th {ACM SIGKDD} {I}nternational
  {C}onference on {K}nowledge {D}iscovery and {D}ata {M}ining}, pages
  1457--1466, 2020.

\bibitem{Buzzicotti2021}
M.~Buzzicotti, F.~Bonaccorso, P.~Clark~Di Leoni, and L.~Biferale.
\newblock Reconstruction of turbulent data with deep generative models for
  semantic inpainting from {TURB-Rot} database.
\newblock {\em Physical Review Fluids}, 6:050503, 2021.

\bibitem{Kim2019}
B.~Kim, V.~C. Azevedo, N.~Thuerey, T.~Kim, M.~Gross, and B.~Solenthaler.
\newblock Deep fluids: Agenerative network for parameterized fluid simulations.
\newblock {\em Computer Graphics Forum}, 38(2):59--70, 2019.

\bibitem{Wu2020}
J.-L. Wu, K.~Kashinath, A.~Albert, D.~Chirila, Prabhat, and H.~Xiao.
\newblock Enforcing statistical constraints in generative adversarial networks
  for modeling chaotic dynamical systems.
\newblock {\em Journal of Computational Physics}, 406:109209, 2020.

\bibitem{Yousif2021}
M.~Z. Yousif, L.~Yu, and H.-C. Lim.
\newblock High-fidelity reconstruction of turbulent flow from spatially limited
  data using enhanced super-resolution generative adversarial network.
\newblock {\em Physics of Fluids}, 33:125119, 2021.

\bibitem{Yousif2022}
M.~Z. Yousif, L.~Yu, and H.-C. Lim.
\newblock Super-resolution reconstruction of turbulent flow fields at various
  {R}eynolds numbers based on generative adversarial networks.
\newblock {\em Physics of Fluids}, 34:015130, 2022.

\bibitem{Yousif2022a}
M.~Yousif, L.~Yu, and H.~Lim.
\newblock Physics-guided deep learning for generating turbulent inflow
  conditions.
\newblock {\em Journal of Fluid Mechanics}, 936:A21, 2022.

\bibitem{GraneroBelinchon2024}
Carlos Granero-Belinchon.
\newblock Neural network based generation of a 1-dimensional stochastic field
  with turbulent velocity statistics.
\newblock {\em Physica D: Nonlinear Phenomena}, 458:133997, 2024.

\bibitem{Richardson1921}
L.~F. Richardson.
\newblock Some measurements of atmospheric turbulence.
\newblock {\em Philosophical Transactions of the Royal Society of London.
  Series A: Containing papers of a mathematical or physical character},
  221:1--28, 1921.

\bibitem{Karman1938}
T.~von K{\'a}rm{\'a}n and L.~Howarth.
\newblock On the statistical theory of isotropic turbulence.
\newblock {\em Proceedings of the Royal Society of London A}, 164:192--215,
  1938.

\bibitem{Chevillard2005}
L.~Chevillard, B.~Castaing, and E.~L{\'e}v{\^e}que.
\newblock On the rapid increase of intermittency in the near-dissipationrange
  of fully developed turbulence.
\newblock {\em The European Physical Journal B}, 45:561--567, 2005.

\bibitem{Tabeling1996}
P.~Tabeling, G.~Zocchi, F.~Belin, J.~Maurer, and H.~Willaime.
\newblock Probability density functions, skewness, and flatness in large
  {R}eynolds number turbulence.
\newblock {\em Phys. Rev. E}, 53:1613--1621, 1996.

\bibitem{Chevillard2010}
L.~Chevillard, R.~Robert, and V.~Vargas.
\newblock A stochastic representation of the local structure of turbulence.
\newblock {\em Europhysics Letters}, 89:54002, 2010.

\bibitem{Moffatt2021}
H.~K. Moffatt.
\newblock Extreme events in turbulent flow.
\newblock {\em Journal of Fluid Mechanics}, 914:F1, 2021.

\bibitem{Kahalerras1998}
H.~Kahalerras, Y.~Malecot, Y.~Gagne, and B.~Castaing.
\newblock Intermittency and {R}eynolds number.
\newblock {\em Physics of Fluids}, 10:910--921, 1998.

\bibitem{GBelinchon2016}
C.~Granero-Belinchon, S.~G. Roux, and N.~B. Garnier.
\newblock Scaling of information in turbulence.
\newblock {\em EuroPhysics Letters}, 115(5):58003, 2016.

\bibitem{Arneodo1999}
A.~Arneodo, S.~Manneville, J.~F. Muzy, and S.~G. Roux.
\newblock Revealing a lognormal cascading process in turbulent velocity
  statistics with wavelet analysis.
\newblock {\em Philosophical transactions: mathematical, physical and
  engineering sciences}, 357(1760):2415--2438, 1999.

\bibitem{Ronneberger2015}
O.~Ronneberger, P.~Fischer, and T.~Brox.
\newblock U-{N}et: convolutional networks for biomedical image segmentation.
\newblock In {\em Medical Image Computing and Computer-Assisted Intervention
  – MICCAI 2015}, pages 234--241, 2015.

\bibitem{Goodfellow-et-al-2016}
I.~Goodfellow, Y.~Bengio, and A.~Courville.
\newblock {\em Deep Learning}.
\newblock MIT Press, 2016.
\newblock \url{http://www.deeplearningbook.org}.

\bibitem{Batchelor1951}
G.~K. Batchelor.
\newblock Pressure fluctuations in isotropic turbulence.
\newblock {\em Mathematical Proceedings of the Cambridge Philosophical
  Society}, 47:359--374, 1951.

\bibitem{Chung1989}
J.~K. Chung, P.~L. Kannappan, C.~T. Ng, and P.~K. Sahoo.
\newblock Measures of distance between probability distributions.
\newblock {\em Journal of Mathematical Analysis and Applications},
  138:280--292, 1989.

\bibitem{Arjovsky2017}
M.~Arjovsky, S.~Chintala, and L.~Bottou.
\newblock Wasserstein {G}enerative {A}dversarial {N}etworks.
\newblock In {\em Proceedings of the 34th {I}nternational {C}onference on
  {M}achine {L}earning}, volume~70, pages 214--223, Sydney, Australia, August
  2017.

\bibitem{Subel2023}
A.~Subel, Y.~Guan, A.~Chattopadhyay, and P.~Hassanzadeh.
\newblock Explaining the physics of transfer learning in data-driven turbulence
  modeling.
\newblock {\em PNAS Nexus}, 2(3):pgad015, 2023.

\bibitem{GraneroBelinchon2019a}
C.~Granero-Belinchon, S.~G. Roux, P.~Abry, and N.~B. Garnier.
\newblock Probing high-order dependencies with information theory.
\newblock {\em IEEE Transactions on Signal Processing}, 67(14):3796--3805,
  2019.

\bibitem{Chen2021}
R.~J. Chen, M.~Y. Lu, T.~Y. Chen, D.~F.~K. Williamson, and F.~Mahmood.
\newblock Synthetic data in machine learning for medicine and healthcare.
\newblock {\em Nature Biomedical Engineering}, 5:493--497, 2021.

\bibitem{Gao2023}
C.~Gao, B.~D. Killeen, Y.~Hu, R.~B. Grupp, R.~H. Taylor, M.~Armand, and
  M.~Unberath.
\newblock Synthetic data accelerates the development of generalizable
  learning-based algorithms for {X}-ray image analysis.
\newblock {\em Nature Machine Intelligence}, 5:294--308, 2023.

\bibitem{Khouider2003}
B.~Khouider, A.~J. Majda, and M.~A. Katsoulakis.
\newblock Coarse-grained stochastic models for tropical convection and climate.
\newblock {\em Proceedings of the National Academy of Sciences},
  100(21):11941--11946, 2003.

\bibitem{Chapron2018}
B.~Chapron, P.~Derian, E.~Memin, and V.~Resseguier.
\newblock Large-scale flows under location uncertainty: a consistent stochastic
  framework.
\newblock {\em Quarterly Journal of the Royal Meteorological Society},
  144(710):251--260, 2018.

\bibitem{Muzy2000}
J.~F. Muzy, J.~Delour, and E.~Bacry.
\newblock Modelling fluctuations of financial time series: from cascade process
  to stochastic volatility model.
\newblock {\em The European Physical Journal B}, 17:537--548, 2000.

\bibitem{Han2016}
J.~Han, X.-P. Zhang, and F.~Wang.
\newblock Gaussian process regression stochastic volatility model for financial
  time series.
\newblock {\em IEEE Journal of Selected Topics in Signal Processing},
  10(6):1015--1028, 2016.

\bibitem{OchabMarcinek2004}
A.~Ochab-Marcinek and E.~Gudowska-Nowak.
\newblock Population growth and control in stochastic models of cancer
  development.
\newblock {\em Physica A}, 343:557--572, 2004.

\bibitem{Baar2016}
M.~Baar, L.~Coquille, H.~Mayer, M.~Holzel, M.~Rogava, T.~Tuting, and A.~Bovier.
\newblock A stochastic model for immunotherapy of cancer.
\newblock {\em Scientific Reports}, 6:24169, 2016.

\bibitem{Schmitt2001}
F.~Schmitt and L.~Seuront.
\newblock Multifractal random walk in copepod behavior.
\newblock {\em Physica A}, 301:375--396, 2001.

\bibitem{Gloaguen2006}
C.~Gloaguen, F.~Fleischer, H.~Schmidt, and V.~Schmidt.
\newblock Fitting of stochastic telecommunication network models via distance
  measures and {M}onte–{C}arlo tests.
\newblock {\em Telecommunication systems}, 31:353--377, 2006.

\bibitem{Eisenblatter2012}
A.~Eisenblatter and J.~Schweiger.
\newblock Multistage stochastic programming in strategic telecommunication
  network planning.
\newblock {\em Computational Management Science}, 9:303--321, 2012.

\end{thebibliography}

\end{document}